# Enhancing Medication Recommendation with LLM Text Representation

Yu-Tzu Lee

## Abstract

Most of the existing medication recommendation models are predicted with only structured data such as medical codes, with the remaining other large amount of unstructured or semi-structured data underutilization. To increase the utilization effectively, we proposed a method of enhancing medication recommendation with Large Language Model (LLM) text representation. LLM harnesses powerful language understanding and generation capabilities, enabling the extraction of information from complex and lengthy unstructured data such as clinical notes which contain complex terminology. This method can be applied to several existing base models we selected and improve medication recommendation performance with the combination representation of text and medical codes experiments on two different datasets. LLM text representation alone can even demonstrate a comparable ability to the medical code representation alone. Overall, this is a general method that can be applied to other models for improved recommendations.

# Table of Content

# List of Figures

# List of Tables

# 1. Introduction

## 1.1. Background

Medication recommendation requires the precise prediction of the drug sets instead of a single item meanwhile preventing the (drug-drug-interaction) DDI risk (Mulyadi and Suk, 2023; Yang et al., 2023). Most of the well-known models for medication recommendation e.g., GAMENet (Shang et al., 2019b), SafeDrug (Yang et al., 2021b), etc. are analyzed based on diagnosis codes, procedure codes, and medication codes, which belong to structured data. However, Electronic Medical Records (EMR) still contain other data such as unstructured and semi-structured data, that is under-utilized in medication recommendation (Bhoi et al., 2021). Unstructured data are the records documented in free text such as discharge summary and admission notes, etc., while semi-structured data includes measurement values in examination reports, lab data, etc.

A study showed that summarized clinical textual data could lead to better decision-making compared with classical graphical visualization (Portet et al., 2009). Some clinical information extraction tools have also been developed to extract events and clinical concepts from text (Aronson, 2001; Wang et al., 2018). These findings indicate that the information extracted from text notes in EMR is under-utilized and still worth investigating. As a result, effectively exploiting and leveraging unstructured medical data has great potential in benefiting healthcare analytics (Adnan et al., 2020a).

However, existing research on unstructured medical data for different tasks, including medication recommendations, was used in earlier practice. For instance, extract symptoms (Sondhi et al., 2012; Tahabi et al., 2023a) using UMLS MetaMap (Aronson, 2001) or predict structured medical codes from unstructured clinical notes through Pre-trained Language Models (PLMs) (Heo et al., 2021). BERT has also been found to underperform in clinical text classification tasks, which may be due to its pretraining and WordPiece tokenization (Gao et al., 2021). Consequently, the disadvantages of existing approaches are primarily as follows: (i) **Relying on preprocessing and pre-training**: Whether solving the problem with PLMs (Huang et al., 2020; Lee et al., 2020) or other tools (Lu et al., 2021), data preprocessing is essential. Additionally, pre-training and fine-tuning with PLMs are more time-consuming. (ii) **Medical domain knowledge required**: Existing works on symptom extraction (Tan et al., 2022) require tools that have external medical knowledge. The pre-training and fine-tuning of PLMs also need notes that contain medical knowledge.

## 1.2. Motivation

Unstructured data, such as clinical text found in medical histories, holds a wealth of undeveloped potential. Despite its richness and the valuable insights that could offer for doctors' decision-making, this data often remains underutilized in medication recommendation. Before the emergence of Large Language Models (LLMs), handling unstructured data dealing with ML or fine-tuning the PLMs required extensive text preprocessing. This procedure was often costly and time-consuming.

In 2020, the introduction of GPT-3 by OpenAI was the major advancement of LLMs development. It was significantly larger than other LLMs at the time having 175 billion parameters (Fan et al., 2023). Afterward, OpenAI announced GPT-4 which is capable of processing both text and image and generating textual outputs. It achieves human-level performance on most of the exams. They can not only accomplish some basic abilities, such as language generation, knowledge utilization, and complex reasoning but also exhibit some superior abilities like human alignment, and interaction with the external environment (Zhao et al., 2023).

Whereas, LLMs have been adopted in various areas to do the work with the human-like thought pattern. For example, replacing the preprocessing or information extraction by LLMs (Agrawal et al., 2022). We can directly obtain responses from LLMs for tasks such as question answering, summarization, or requesting advice. LLMs excel over PLMs in this aspect without the need for fine-tuning. Discharge summary like clinical notes in our datasets are such problems of processing the textual data, which is suitably resolved with LLMs. Additionally, LLMs can execute text mining or knowledge extraction from medical notes in a manner distinct from traditional ML and PLMs.

Take the mentioned NER task for example, it has been shown to enhance performance through prompt engineering by LLMs like GPT (Hu et al., 2024). Additionally, the implementation of a more refined prompt design for LLMs has led to structured output texts, resulting in enhanced retrieval of information like medication from medical history (Agrawal et al., 2022). To our best knowledge, the application of LLM hidden states for medication recommendation has so far been limited to input with structured medical codes (Liu et al., 2024) and has not been employed with unstructured clinical notes.

Therefore, we anticipate that medication recommendation can make progress with text representation from LLM's language understanding abilities. This approach considers not only using medical codes but also incorporating information from clinical notes to further improve predictions.

## 1.3. Objectives

After reviewing the shortage of existing research on using medical codes only or the redundant processing of clinical notes. We design experiments to apply LLMs for text representation extraction on two datasets: the well-known medical datasets MIMI-III and hospitalized data from Ditmanson Medical Foundation Chia-Yi Christian Hospital (CYCH). The output text representation from LLMs is combined with the medical codes embedding, resulting in embedding that incorporates both aggregated structured and unstructured data.

We aim to leverage off-the-shelf LLMs, which have been trained for general purposes and are adept at answering questions across various domains, except for some specific knowledge they may not have acquired. Capitalizing on this untapped information through LLM's ability to understand, we intend to add extra underused information after being digested by LLMs. That is simulating how a doctor prescribes medication based on reviewing clinical records to enhance the drug recommendation.

## 1.4. Thesis Organization

The organization of this study is as follows. In Chapter 1, the background and context of the study are introduced, highlighting the importance of the research, and outlining the primary objectives and the structure of the thesis. Chapter 2 presents a detailed review of previous methodologies for medication recommendation tasks, covering structured data analysis, unstructured medical data, large language models (LLMs), and their applications and limitations, as well as the specific issues addressed by this study. Building upon the comprehensive discussion in Chapter 2, the methodology for this study is introduced in Chapter 3. This includes our method, definitions of key terms, descriptions of the datasets used, data preprocessing steps, base models, implementation settings, and evaluation metrics. Chapter 4 analyzes the results of the experiments, focusing on the effectiveness of LLM text representation and the combined representation of text and medical codes. Finally, Chapter 5 encapsulates the key findings, discusses the contributions and limitations of the study, and proposes future research directions based on the insights gained from this research.

# 2. Related Works

## 2.1. Medication Recommendation with Medical Structured Data

According to Shang et al. (2019b), instance-based and longitudinal are the two main approaches used earlier in Medication combination recommendation. Instance-based methods provide recommendations only after analyzing the current visit records. Leap (Zhang et al., 2017) has implemented multi-instance multi-label learning to reference multiple diagnosis codes of each visit. However, the drawback of the instance-based method is its inability to refer to the same patient's medical history (Liao, 2023), and longitudinal methods are the answer to this shortcoming.

Longitudinal methods consider the entire sequence of patient visits over time. MICRON (Yang et al., 2021a) and COGNet (Wu et al., 2022) emphasized the changes in the patient's states. Particularly, COGNet, as shown in Figure 1, proposed a "copy-or-predict" mechanism to determine whether to use the medication from previous recommendations or make a new prediction. This imitates a real doctor's prescription and could capture the medication changes more effectively. It also introduced the Transformer architecture to train with Encoder-Decoder. DCw-MANN (Le et al., 2018), GAMENet (Shang et al., 2019b), AMANet (He et al., 2020), and ARMR (Wang et al., 2021) DAPSNet (Wu et al., 2023a) employed the memory augmented networks. SHAPE (Liu et al., 2023), as shown in Figure 2, designed their framework with two encoders: one for intra-visit, which operates at the visit-level, and one for inter-visit, which operates at the patient-level longitudinal. Moreover, they proposed an adaptive curriculum manager to dynamically assign the complex coefficients for each patient, addressing the challenge of predicting medication for short visit records that lack historical medication information. Overall, discarding patient data with only one visit record is the disadvantage of longitudinal methods.

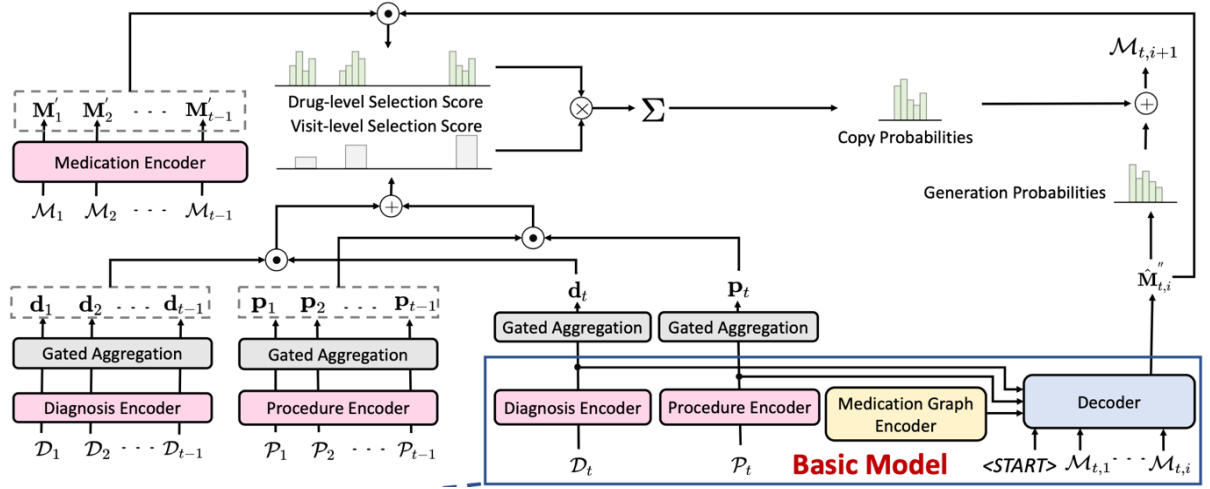

Figure 1 COGNet architecture (Wu et al., 2022)

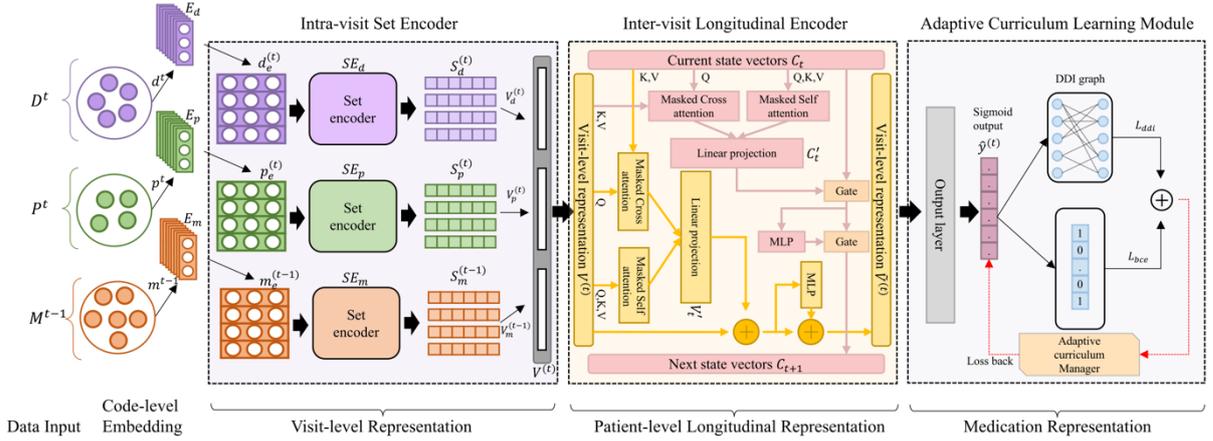

Figure 2 SHAPE architecture (Liu et al., 2023)

Further, recent advancements have introduced Graph Neural Network-based (GNN-based) models to link the embedding as memory networks. G-BERT (Shang et al., 2019a), GAMENet (Shang et al., 2019b), SafeDrug (Yang et al., 2021b), PREMIER (Bhoi et al., 2021), COGNet, MoleRec (Yang et al., 2023), GSVEMEed (Liao, 2023), and KindMed (Mulyadi and Suk, 2023). have leveraged GNNs in longitudinal methods, resulting in enhanced performance. Among them, GAMENet and MoleRec, along with the longitudinal models COGNet and SHAPE, utilized GNN to learn the information of DDI, while SafeDrug and PREMIER used GNN to learn about the drug's molecular. GAMENet, as illustrated in Figure 3, incorporated external

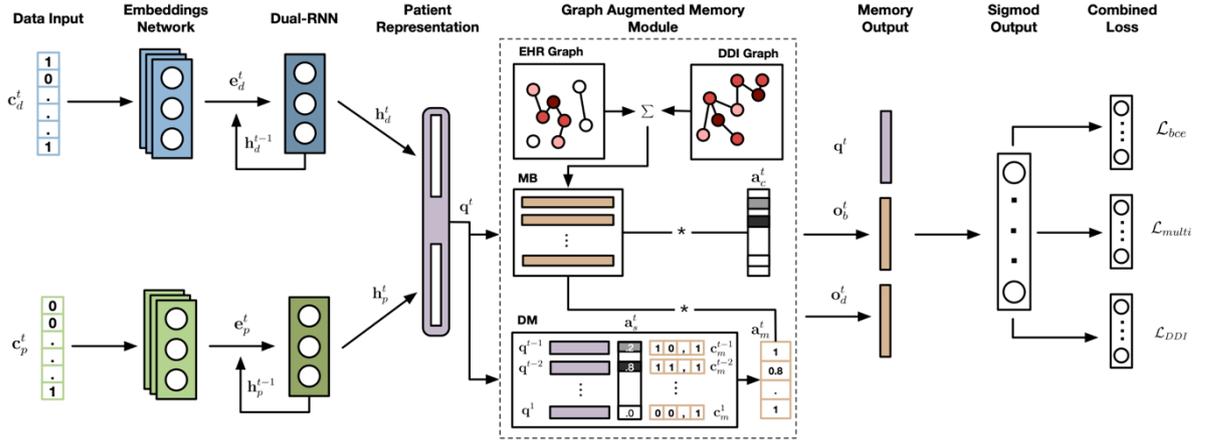

Figure 3 GAMENet architecture (Shang et al., 2019b)

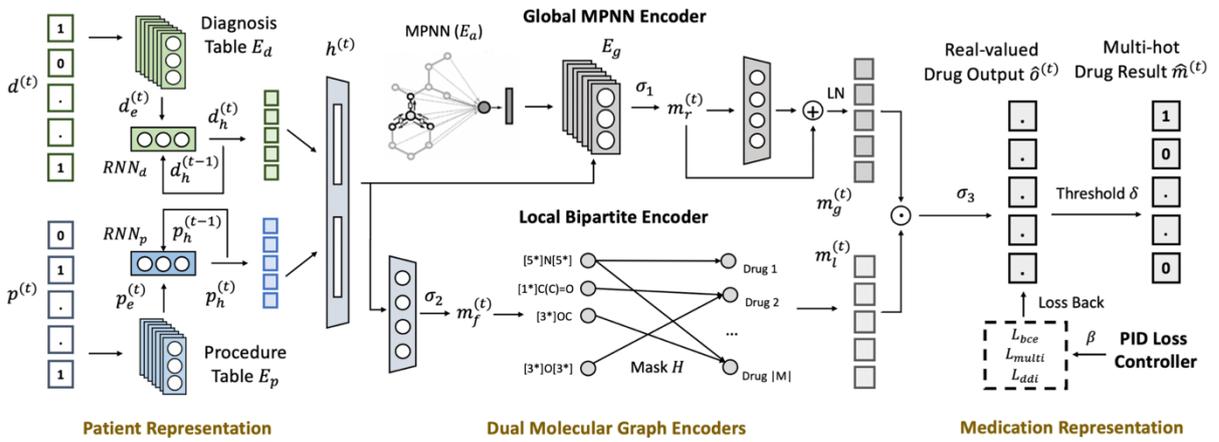

Figure 4 SafeDrug architecture (Yang et al., 2021b)

DDI knowledge and integrated multiple modules building both EHR and DDI graphs using GNN. SafeDrug, shown in Figure 4, was the first to incorporate drug molecular structures through GNN in both global and local information further enhancing the model with the DDI knowledge. Considering DDI in loss computation distinguished it from the approach of GAMENet. G-BERT, as shown in Figure 5, was the first to adopt a language model in medication recommendation and include the records of patients with only one visit in the EMR during the pre-training, addressing the main disadvantage of longitudinal methods. They utilized GNN to build ontology trees by integrating the medical codes' ancestor information, obtaining ontology embedding. KindMed employed GNN as the medical knowledge graph and devised two additional modules: a fusion module to merge the historical features with deep fusion, and an attentive prescribing module for encoding the temporal dynamics that consider the patient's historical admission. GSVEMEed, shown in Figure 6, was the first to utilize GNN for referencing visit history across patients and incorporating records of patients with only one visit into the entire training stage. This differs from G-BERT, which used records of patients with only one visit solely for pre-training.

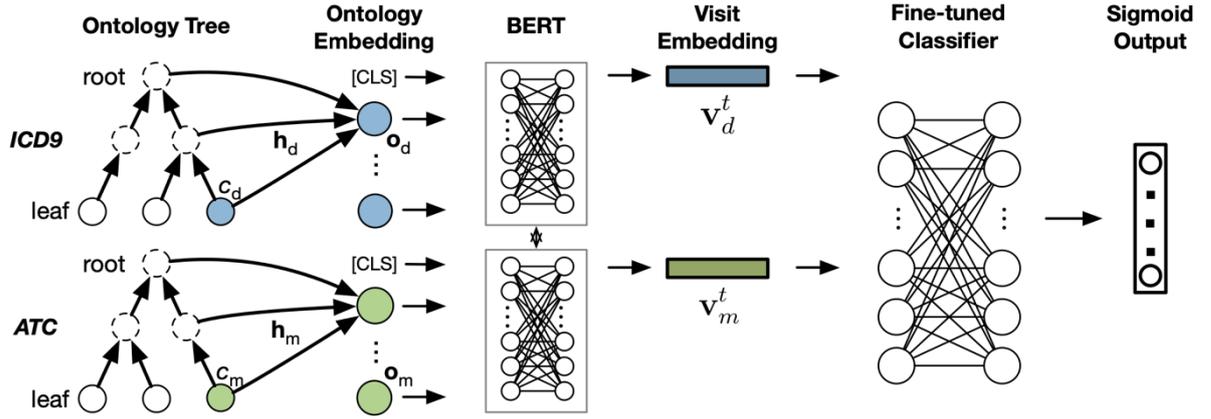

Figure 5 G-BERT architecture (Shang et al., 2019a)

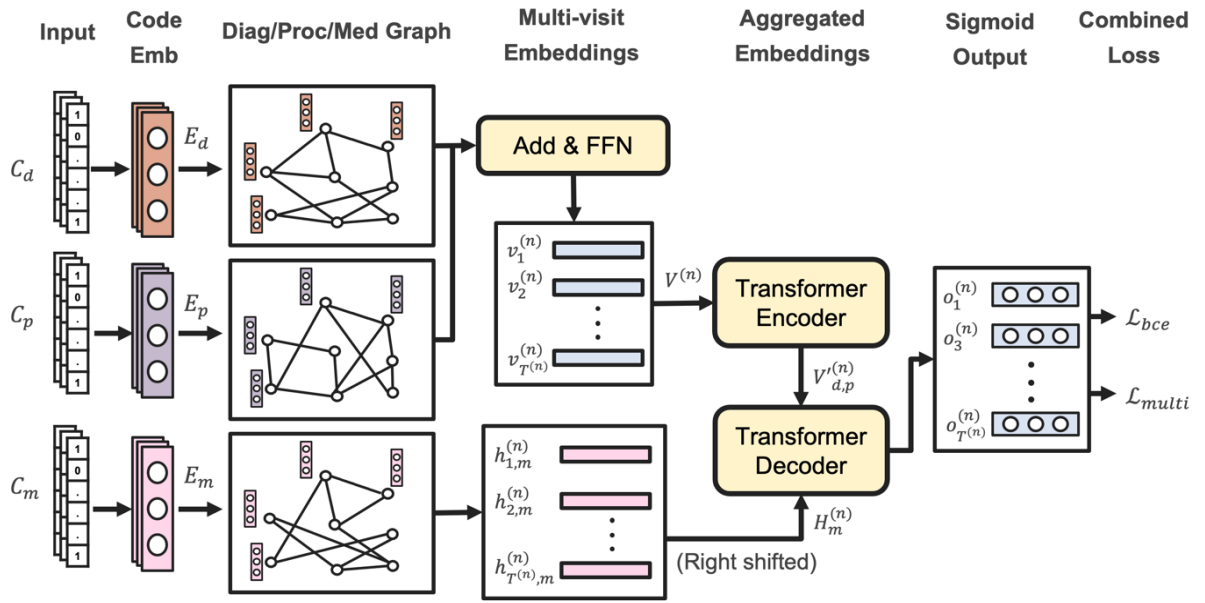

Figure 6 GSVEMEed architecture (Liao, 2023)

MEGACare (Wu et al., 2023b) and StratMed (Xiang Li et al., 2024) employ GNN to support more complex frameworks. MEGACare proposed capturing the high-order correlation between patent visits and medical codes for better patient representation by using a hypergraph structure with multi-view. MEGACare utilized GNN for the code graph encoding and EHR hypergraph construction and was capable of predicting both diagnosis and medication simultaneously. StratMed, as shown in Figure 7, emphasized grouping the relationship of the low-frequency data to increase their importance, helping the model to learn sparse data. Meanwhile, it shared SafeDrug's concern on safety, maintaining performance in both safety and accuracy through dual-property representation constructed by GNN at the visit level.

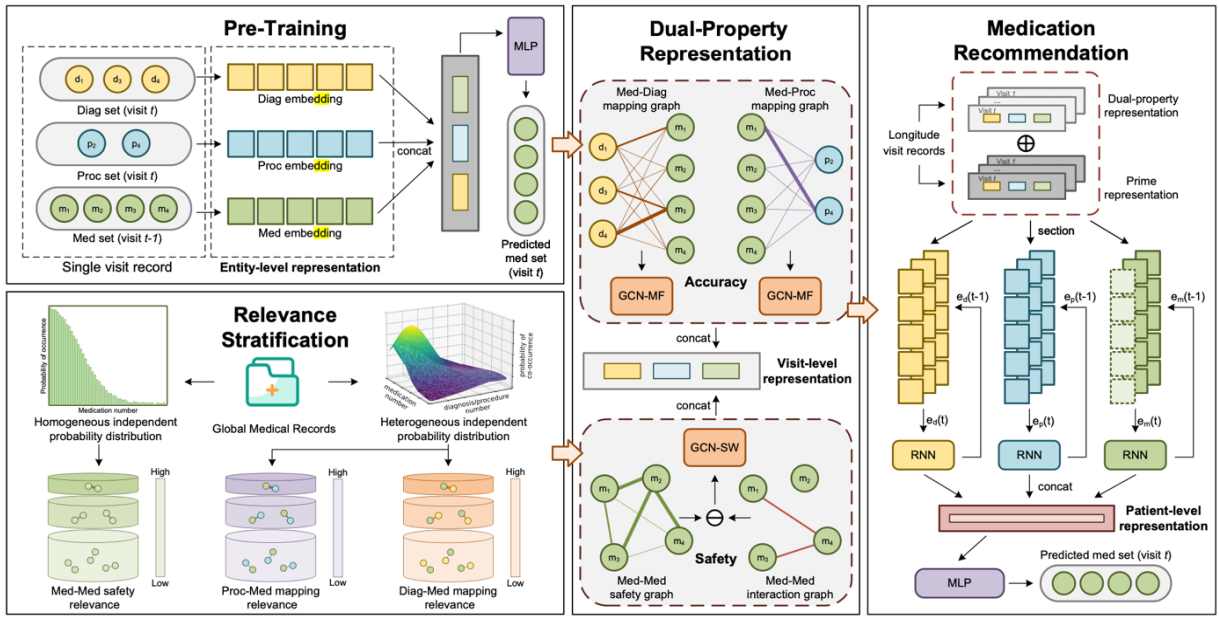

Figure 7 StratMed architecture (Xiang Li et al., 2024)

All the mentioned approaches above primarily focus on processing the experiments with structured data such as diagnosis codes, procedure codes, and medication codes, though there are still unstructured data and semi-structured data in EMR leading to underutilized. The overall comparison of each method can be referred to Table 1.

Table 1 Related Study of Medication Recommendation with Medical Structured Data

| Method | Model | Code | Properties | Weakness |
|---|---|---|---|---|
| **Instance-based** | Leap (Zhang et al., 2017) | - | Reference multiple diagnosis codes of each visit through multi-instance multi-label learning. | Unable to refer to the same patient's medical history. |
| **Longitudinal** | COGNet (Wu et al., 2022) | https://github.com/BarryRun/COGNet | Capture the patient's medication changes by considering medication history and the copy-or-predict mechanism. | Discarding patient data with only one visit record. |
| | SHAPE (Liu et al., 2023) | https://github.com/sherry6247/SHAPE/tree/main | Encode the embeddings with two Encoders: One for visit-level representation and one for patient-level longitudinal representation. They also proposed an adaptive curriculum manager to address the challenge of predicting medication for short visit records that lack historical information. | |
| **Graph-based (GNN)** | G-BERT (Shang et al., 2019a) | https://github.com/jshang123/G-Bert | The first to adopt a language model in medication recommendation and include the records of patients with only one visit during the pre-training. GNN was used to build ontology trees by integrating the medical codes' ancestor information, obtaining ontology embedding. | Only focusing on the processing of structured data. |
| | GAMENet (Shang et al., 2019b) | https://github.com/sjy1203/GAMENet | Incorporated external DDI knowledge and integrated multiple modules building both EHR and DDI graphs using GNN. | |
| | SafeDrug (Yang et al., 2021b) | https://github.com/ycq091044/SafeDrug | The first to incorporate drug molecular structures through GNN in both global and local information further considering DDI in loss computation. | |

| GSVEMEed (Liao, 2023) | - | The first to utilize GNN for referencing visit history across patients and incorporating records of patients with only one visit into the entire training stage. |
| KindMed (Mulyadi and Suk, 2023) | - | Employed GNN as the medical knowledge graph and devised two additional modules: a fusion module to merge the historical features, and an attentive prescribing module for encoding the temporal dynamics that consider the patient's historical admission. |
| MEGACare (Wu et al., 2023b) | https://github.com/senticnet/MEGACare | Capturing the high-order correlation between patent visits and medical codes for better patient representation by using a hypergraph structure through GNN with multi-view. |
| StratMed (Xiang Li et al., 2024) | https://github.com/lixiang-222/StratMed/tree/main | Emphasized grouping the relationship of the low-frequency data to increase their importance, maintaining performance in both safety and accuracy through dual-property representation constructed by GNN at the visit level. |

## 2.2. Analysis of Unstructured Medical Data

Processed unstructured medical data can be applied to various tasks, including medication recommendation. Extracting symptoms from clinical records is the practice in most ancillary medication recommendation studies. 4SDrug (Tan et al., 2022) was proposed to predict drug sets according to the coordinate symptom sets using set-oriented representation. Another drug recommendation causal graphical model (Sun et al., 2022) encodes the procedure contexts, diagnosis contexts as well as the extracted symptoms contexts from clinical notes into embeddings.

Several studies utilize unstructured medical data for different tasks, commonly employing UMLS MetaMap (Aronson, 2001) to extract symptoms from clinical notes and construct a symptom graph. For instance, SympGrap (Sondhi et al., 2012) is built for disease diagnosis prediction, and SymptomGraph (Tahabi et al., 2023b) is designed to identify the symptom clusters. Additionally, MedText (Lu et al., 2021) and DeepNote-GNN (Golmaei and Luo, 2021) are predicting patient readmission risk. Furthermore, Buckland et al. (2021) regarded the extracted symptoms as features for classifying suicide attempts. Among these approaches, the extracted symptoms can be used to construct a graph or encoded directly as embeddings.

Transforming the unstructured data into structured form could be a crucial task in the medical field, since admission notes, clinical notes, or discharge summary are associated with a patient's history (Nuthakki et al., 2019). Consequently, some studies focus on predicting medical codes from discharge summary (Heo et al., 2021). This can also be viewed as a multilabel classification task, encompassing diagnosis codes, diagnosis categories, and procedure codes. Nonetheless, Gao et al. (2021) have shown that the BERT-based model is worse at clinical text classification.

Clinical note summarization is another important area of research. SPeC (Chuang et al., 2024) proposed optimizing the soft prompt to improve the summarization performance of

LLMs.

van Aken et al. (2021) proposed a novel admission-to-discharge task, which involves predicting four clinical outcomes, including diagnoses at discharge, procedures performed, in-hospital mortality, and length-of-stay prediction from MIMIC-III discharge summary. Admission notes have to be split from the discharge summary first.

Some studies have pre-trained or fine-tuned the models using data with medical knowledge. For instance, ClinicalBERT (Huang et al., 2020) was pre-trained on clinical notes and fine-tuned on readmission prediction task. BioBERT (Lee et al., 2020) pre-trained on Wikipedia and BooksCorpus and fine-tuned on NER, RE, and QA tasks.

Overall, the majority of unstructured medical data processing involves collating it into structured formats, with only a limited amount being embedded in terms extracted from clinical notes (Sheikhalishahi et al., 2019). Procedures for transforming unstructured data into structured include Named-Entity Recognition (NER) and Relation Extraction (RE) (Landolsi et al., 2023). The variety and the vast volume of unstructured data leading numerous technical challenges, including integration, preprocessing, information extraction, etc. (Adnan et al., 2020b). There are two main categories of methods for clinical information extraction, rule-based and Machine Learning (ML) (Wang et al., 2018). While processing in either way, a main challenge in clinical Natural Language Processing (NLP) is information overload which could lead to a substantial obstacle to accessing crucial information within vast datasets. Moreover, semantic and context understanding are essential aspects of NLP. The diversity of text formats also poses a significant challenge (Hossain et al., 2023).

## 2.3. Large Language Models (LLMs)

Large Language Models (LLMs) are large-scale pre-trained language models, distinguished from Pre-trained Language Models (PLMs) (Minaee et al., 2024). PLMs are trained on a large amount of data that learn a general understanding of language, including vocabulary, syntax, and semantics. After pre-training, PLMs can be fine-tuned for specific tasks, such as sentiment analysis, question answering, machine translation, etc. The fine-tuning process could adapt the model to the specific data and requirements of the task.

However, LLMs often contain tens and hundreds of billions or even trillions of parameters that they have better language understanding and generation abilities, enabling them to handle more complex tasks. Furthermore, LLMs have emergent abilities (Wei et al., 2022) for example, in-context learning, instruction tuning, and multi-step reasoning. Hence, in terms of versatility, LLMs could have a wide range of applications and be more powerful than PLMs.

Most LLMs are based on the Transformer architecture (Vaswani et al., 2017), including models like GPT-4 by OpenAI (OpenAI, 2023), Llama by Meta (Touvron et al., 2023), Mistral 7B (Jiang et al., 2023), and PaLM by Google (Chowdhery et al., 2023), etc. Nevertheless, there are still some non-transformer LLMs that have been proposed (Minaee et al., 2024). Among these LLMs, some are open-source, and publicly available, while others are closed-source. For instance, GPT-3 (Brown et al., 2020) and GPT-4 are not publicly available and are subject to fees. In contrast, all models in the Llama family are not only publicly available but also free. Therefore, open-source models like Llama can be applied more freely. Enabling both the operation of the model as well as access to its weights would enhance the feasibility of research for value creation.

Recently, many studies have leveraged the power of LLMs in various fields, such as medical, education, finance, engineering, marketing, etc. (Hadi et al., 2023). Consequently, LLMs are possibly very helpful for general-purpose use. In this research, we primarily utilize the open-source Mistral 7B-Instruct, which has been fine-tuned on instruction datasets, for our experiment.

## 2.4. LLM Application to Medical Unstructured Data

In Chapter 2.2, we have explored how unstructured medical data can be analyzed. In this section, we will discuss the adoption of LLMs in the medical domain. Most of the research in Chapter 2.2 focuses on converting unstructured to structured data, including extracting symptoms. These are the parts of information extraction. A study (Agrawal et al., 2022) implemented LLMs to enhance information extraction, specifically to predict structured outputs with LLM. Also, GPT-3.5 and GPT-4 were applied to the clinical NER tasks (Hu et al., 2024) to improve the performance via prompt engineering.

Some studies have developed LLM for the clinical domain. For instance, GatorTron (Yang et al., 2022) is a large clinical language model built from scratch for EMR. Google has also trained an LLM for clinical through instruction prompt tuning. The resulting model is called Med-PaLM (Singhal et al., 2023), which currently requires running on Google Cloud Platform and is not supported in Taiwan at the time of writing this thesis.

## 2.5. Knowledge in LLMs

LLMs possess extensive knowledge because of learning vast amounts of information. Leveraging the strengths derived from LLM knowledge enables us to generate value. Nevertheless, LLMs are still a "black box" in terms of how they interpret information. Hidden states are the middle output of the LLMs, which contain the LLM's understanding of the learned information knowledge.

Some studies have engaged in knowledge extraction from LLMs. "Knowledge Extraction" may have different definitions depending on the domain. Such as in an information extraction-related context (Martinez-Rodriguez et al., 2020), it differs from the neural networks we discuss here.

### 2.5.1. Applications for Extracted Knowledge from LLMs

The hidden states in LLMs have been used for personalized news recommendation (Hao et al., 2023). News representations generated by LLM contain rich semantic information. These representations are then combined with the information about news entities generated by Knowledge Graphs (KG). Their result showed that the integration of LLM and KG can achieve more accurate personalized news recommendations. The LLMs they used include ChatGLM2 (Zeng et al., 2023), Llama 2, and RWKV (Peng et al., 2023).

In a different domain, LLM is employed to address multimodal problems for image retrieval, novel image generation, and multimodal dialogue (Koh et al., 2023). They integrated LLM with a text-to-image generation model. The trained model accepts input in the form of either text or images, yet both types of input are simultaneously processed by the LLM during training. First, the hidden representations of text would be accessed from the last hidden layer of LLM. The output embeddings include generated text tokens and special [IMG] tokens representing the image part. After that, a mapping network would translate these hidden representations into the embedding space of a visual model for image retrieval or generation. OPT-6.7B (Zhang et al., 2022) is employed as their model, producing hidden states with 4096 embedding dimensions.

Zhu et al. (2024) employed LLaMA-2-7B, LLaMA-2-13B, and Mistral-7B to acquire their hidden states and aimed to solve math problems. They conducted experiments to identify the original numbers by linear probing the obtained hidden states. Their analysis of LLM's compression numbers demonstrated that intermediate layers exhibited the best performance. However, Mistral-7B was able to maintain the precision in deeper layers which LLaMA-2 could not achieve.

Pal et al. (2023) proposed predicting the model's probability distribution of the next several tokens by using a single hidden state from a specific layer of one token in the Transformer. This suggested that the representation contains such information. Moreover, they observed a phenomenon where predicting the direct next-token resulted in different performance compared to predicting subsequent-token. The best accuracy appeared at the middle-layer hidden states when predicting the direct next-token, while the best accuracy was obtained at the last layer when predicting the subsequent-token.

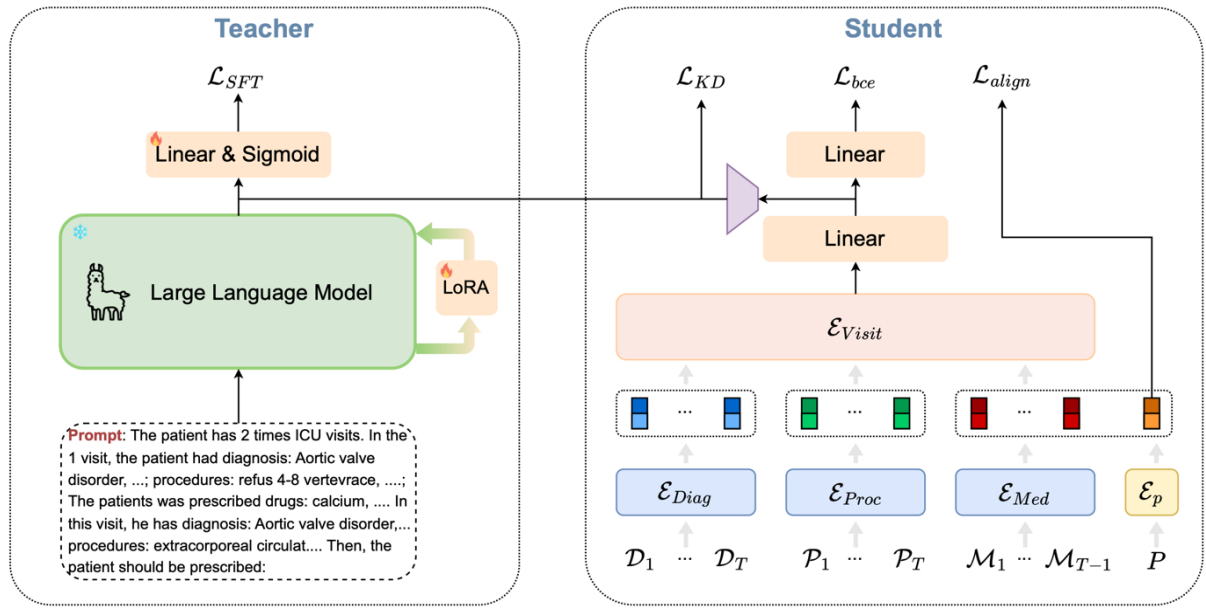

Figure 8 LEADER architecture (Liu et al., 2024)

Azaria and Mitchell (2023) adopted OPT-6.7b and LLAMA2-7b to ask the LLMs to determine whether the statements were true or false. They assumed the values in the hidden states incorporated the information for distinguishing the truthfulness of the statements. The classification results demonstrated the best performance of OPT-6.7B and LLAMA2-7B occurred in different layers. Precisely, OPT-6.7B excelled in the 20th layer, while LLAMA2-7B was optimal in the 16th layer.

LEADER (Liu et al., 2024) as shown in Figure 8 applied LLM hidden states in the medication recommendation. They employed LLM to perform knowledge distillation using diagnosis, procedure, and medication codes. They designed a prompt template (Figure 9) and filled in the medical terms of the three kinds of codes for each visit to fine-tune LLM as a teacher model. Then they took the hidden states from the last layer of LLM to train a smaller student model. Instead of generating word tokens, they replaced the output layer of the LLM with a linear and sigmoid function to obtain the medication probability.

> **Input Prompt Template**
>
> The patient has <VISIT_NUM> times ICU visits.
> In the 1 visit, the patient had diagnosis: <DIAG_NAME>, ..., <DIAG_NAME>; procedures: <PROC_NAME>, ..., <PROC_NAME>; The patient was prescribed drugs: <MED_NAME>, ..., <MED_NAME>. In the 2 visit, ....
> In this visit, the patient has diagnosis: <DIAG_NAME>, ..., <DIAG_NAME>; procedures: <PROC_NAME>, ..., <PROC_NAME>. Then, the patient should be prescribed:

Figure 9 LEADER prompt templates (Liu et al., 2024)

These studies suggest that the knowledge extracted from LLMs is potentially helpful in retrieving useful information.

### 2.5.2. Edit Knowledge in LLMs

Several studies have analyzed the knowledge within the Transformer-based language model, incorporating weight-modified, weight-preserved, meta-learning methods, and optimization-based methods as categorized by Li et al. (2024). However, we are focusing on how LLM's knowledge, namely hidden states, has been analyzed. Li et al. (2024) noted weight-preserved including introducing external models (Mitchell et al., 2022) or altering the LLMs' representation space (Hernandez et al., 2023).

For optimization-based methods, Geva et al. (2021) noted that Feed-forward layers hold the majority of parameters of the Transformer model and demonstrated that feed-forward layers function as key-value memories. They showed that the keys resemble human-interpretable input sequences with certain patterns, while the values generated distributions over the output vocabulary similar to the distribution of predicted next-token from the corresponding key. Then the model output is shaped by aggregating these distributions.

As for weight-modified approaches, Meng et al. (2022a) had a different point of view from Geva et al. (2021) mentioned in the former. They hypothesized MLPs could be a linear associative memory and designed a new intervention method called Rank-One Model Editing (ROME). It can edit one MLP layer using ROME by replacing the knowledge and performing its generalization and specificity, compared with baselines, including the fine-tuning method. They also find mid-layer MLP layers at the final subject token performance peaks. A method called MEMIT (Meng et al., 2022b) editing the memories by Updating the specific layer parameters in Transformer. Experiments were conducted with two LLMs GPT-J (6B) and GPT-NeoX (20B) by comparing MEMIT with a fine-tuning approach, a hypernetwork-based model MEND (Mitchell et al., 2021), and a model editing method ROME (Meng et al., 2022a). The results showed that MEMIT can improve specificity, generalization, and fluency.

Concerning weight-preserved, the knowledge editor REMEDI (Hernandez et al., 2023) alternated the model output by modifying the hidden representation, subsequently generating the desired text.

A model editing technique PMET (Xiaopeng Li et al., 2024) optimized the hidden states of the component in the Transformer layer, namely Multi-Head Self-Attention (MHSA) and Feed-Forward Network (FFN), while only updating FFN weights with the optimized hidden states of FFN. They stated that this differed from the assumptions mentioned above. Similarly, they compared the fine-tuning, MEND, ROME, and MEMIT.

These studies explored how LLM's knowledge could be manipulated, whether by editing weights or hidden representations. Revealing such information is inside LLMs.

## 2.6. Discussion

Medication recommendations primarily rely on structured data, essentially medical codes,

as discussed in Section 2.1 disregarding unstructured data. Nevertheless, research examining unstructured medical data in Section 2.2 has been adopted for multiple tasks, with little focus on medication recommendations. Besides, the processing of unstructured medical data mainly followed conventional methods, involving NLP procedures such as the preprocessing of clinical notes, which incurs significant time and cost.

LLMs possess remarkable capabilities in language understanding, processing, and generation, making them well-suited for handling vast amounts of unstructured medical data. We have learned that the knowledge of how LLMs understand the input is encoded within the hidden states, as demonstrated in Section 2.5. Furthermore, hidden states have been applied to solve some tasks. However, to our best knowledge, the hidden states have not been employed in the unstructured clinical notes. Only LEADER has utilized LLM hidden states to train a model for medication recommendation through knowledge distillation, but they still used only medical codes as input instead of clinical notes. Most applications of LLMs in the medical field have focused on enhancing information extraction or training LLMs with clinical knowledge, as depicted in Section 2.4. Thus, we suggest deploying LLM hidden states for medication recommendation in this study. Additionally, as noted in Section 2.5.1, the best performance may appear in different hidden layers when using different LLMs. Hence, it is important to identify the appropriate hidden layer according to the LLM we choose.

# 3. Methodology

We propose a method of enhancing medication recommendation with LLM text representation. The framework of our method is shown in Figure 10 and can be applied to the several existing base models we selected.

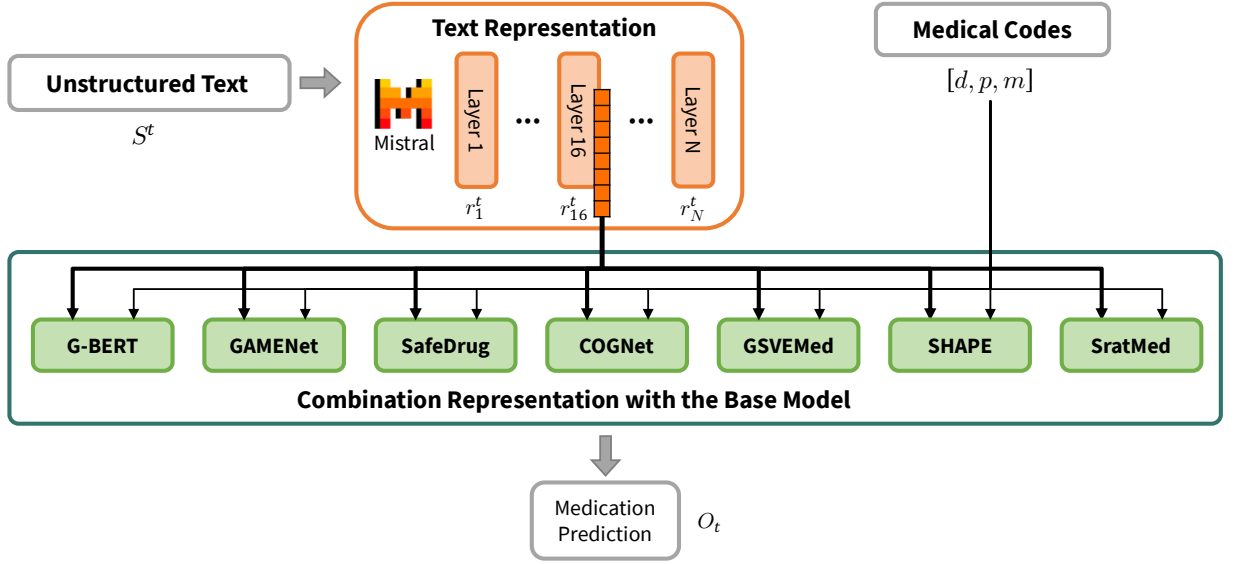

Figure 10 The framework of enhancing medication recommendation with LLM text representation. It mainly contains two parts: Text Representation Extraction and Combination Representation with the Base Model. The text representation $r_i^t$ is first extracted from LLM hidden states, where we set $i = 16$ to denote layer 16. Then, the text representation is combined with medical code embeddings following the procedure defined by the base model to make medication predictions.

## 3.1. Proposed Method

First, given the unstructured clinical notes $S^t$ of a single visit $t$, the text representations $r^t$ are extracted from LLM hidden states. For, structured medical codes $[d, p, m]$, the processing procedure depends on the base models, resulting in medical code embeddings $[e_d, e_p, e_m]$ for diagnosis, procedure, and medication, respectively.

We supposed that the extracted text representation $r^t$ from LLM hidden states could be combined appropriately with the medical codes embeddings. This indicates the synthesis of the structured and the unstructured data. The combination operation of the representations $d \oplus p \oplus$ $r$ also follows the architecture of the base models, hence the notation $\oplus$ represents either addition or concatenation.

The base models then take the combination representations as input and make medication predictions for visit $t$. Evaluate the loss between predicted medication $O_t = [o_1, o_2, ... o_T]$ and the ground true $m$.

### 3.1.1. LLM Text Representation Extraction

We choose Mistral-7B-Instruct-v0.2, an instruction fine-tuned version of Mistral 7B, as our text representation extractor. Since we want to extract the hidden from LLM, we need the open-source LLM. Mistral-7B-Instruct (Jiang et al., 2023) has shown results superior to other open-source LLMs such as Llama 2 13B Chat. Despite having only a 7B model size, it outperforms the chat capabilities of the 13B model size.

Clinical notes in both MIMIC-III and CYCH are typically long. Although the context window size of Mistral-7B-Instruct-v0.2 is 32,000, which is longer than that of Mistral-7B-Instruct-v0.1, our device is still limited in processing complete clinical texts of a single visit. Inspired by Golmaei and Luo (2021), Chang et al. (2023), and Goel et al. (2023), clinical notes $S^t$ are split into several chunks $s_k = \left[s_1, s_2, \ldots, s_{\lceil \frac{L}{K} \rceil}\right]$, where $L$ is the sequence length of $S^t$ and $K < D$ is the length of a single chunk, which is smaller than the context window size $D$. In addition, the clinical notes of CYCH are mostly in English but with some Chinese interspersed, Mistral can still process Chinese input and generate responses in English.

As shown in Figure 11, the text representation $r_i^t$ of clinical notes $S^t$ from visit $t$ using Mistral-7B-Instruct-v0.2 is the average of the hidden states $r_i$ of the input sequence, as described in Equation 1.

$$r_i^t = \frac{1}{L}\left[r_i^{(s_1)} \parallel r_i^{(s_2)} \parallel \ldots \parallel r_i^{\left(s_{\lceil \frac{L}{K} \rceil}\right)}\right] \qquad \text{Equation 1}$$

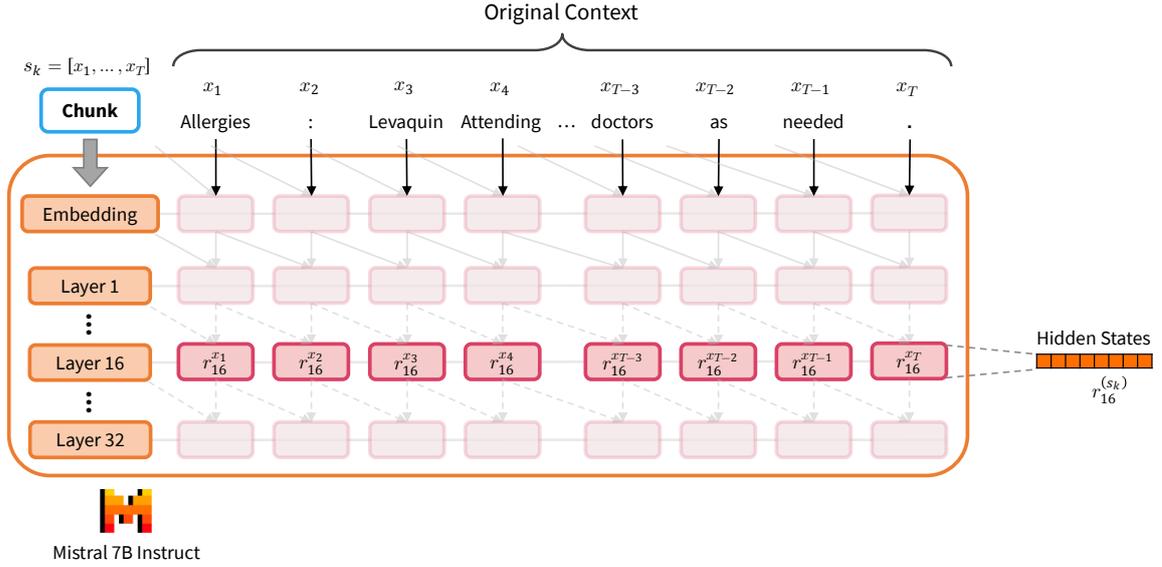

Figure 11 LLM text representation extraction from a chunk. The input for LLM each time is the chunk. We only take the hidden states of layer 16. The hidden states of each chunk are concatenated to the original sequence length of clinical notes and then averaged.

Before averaging, the hidden states $r_i^{(s_k)}$ of the chunks $s_k$ are concatenated into the original sequence length $L$ first. Here we reference the result of Azaria and Mitchell (2023) and Zhu et al. (2024) to set $i = 16$ as the intermediate layer.

### 3.1.2. Combination Representation of Text and Medical Codes

After text representation extraction, it is sent to the selected base models. Here we use GSVEMed as an example to explain the medical code embedding generation and the combination operation, as shown in Figure 12; the detailed operations of other base models can be found in Appendix A.

*3.1.2.1. GSVEMed with LLM Text Representation - Overview*

Patient records in EMR comprise both structured and unstructured data. Patients are denoted as $P^n = \left[ x_1^{(n)}, x_2^{(n)}, \ldots, x_{T^{(n)}}^{(n)} \right]$, where $x^{(n)}$ represents a single visit of patient $n$. For patient $n$, the medical records in EMR are $x_n = (x_n^1, x_n^2, \ldots, x_n^T)$, where $T$ is the total number of visits of patient $n$.

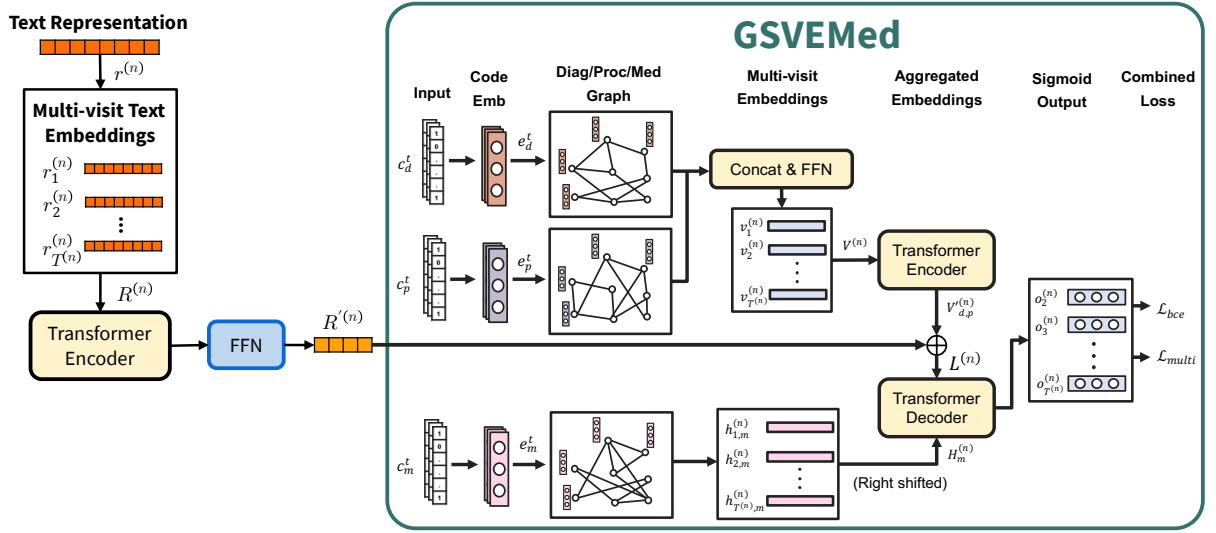

Figure 12 Architecture of GSVEMed with LLM Text Representation. On the left side is the text representation generation, and on the right side is the original GSVEMed model.

For structured data, every single visit $x_n^t = (c_d^t, c_p^t, c_m^t)$, where $t \in \{1, 2, \ldots, T\}$, $c_d^t$ represents diagnosis codes, $c_p^t$ represents procedure codes, and $c_m^t$ represents medication codes. $c_* \in \mathbb{R}^{|\mathcal{C}_*|}$ is the multi-hot vector of the category of $*$. $\mathcal{C}_*$ notes as the medical codes set of categories $*$. Only diagnosis codes and procedure codes are input to the encoder, while medication codes are shifted-right before being processed by the Transformer Decoder. Decoder's output is passed through a Sigmoid function to get the predicted probability for each medication in each visit.

In graph visit representation learning, $\mathcal{G}_d = \{\mathcal{V}_d, \mathcal{E}_d\}$, $\mathcal{G}_p = \{\mathcal{V}_p, \mathcal{E}_p\}$, $\mathcal{G}_m = \{\mathcal{V}_m, \mathcal{E}_m\}$ are the visits graphs of diagnosis codes, procedure codes, and medication codes, respectively, where $\mathcal{V}_* = \{c_*^1, c_*^2, \ldots, c_*^T\}$ is the multi-hot vectors of $c_*$ and $\mathcal{E}_*$ is the edge of among all visits. $A_* \in \mathbb{R}^{T \times T}$ represents the adjacency matrices of the graphs. If the Jaccard Similarity score between visit $t$ and target node $j$ is among the top $K$ highest among all visits with admission times earlier than $j$, then $A[t, j] = 1$. Additionally, to ensure that only the latter visits can reference earlier ones, $t$ must be less than or equal to $j$. $A_{DDI} \in \mathbb{R}^{|\mathcal{C}_m| \times |\mathcal{C}_m|}$ denotes the adjacency matrices of the DDI rate defined according Tatonetti et al. (2012). If the selected

pairwise medications $i$ and $j$ have DDIs, $A_{DDI}[i,j] = 1$; otherwise it is 0.

*3.1.2.2. GSVEMed - Medical Codes Embedding*

There are three kinds of medical codes in our datasets, namely procedure codes, diagnosis codes, and medication codes. Each type of medical code has its embedding table $W_{*,e} \in \mathbb{R}^{|\mathcal{C}_*| \times d}$ to embed the corresponding medical code $c_*^t$ within each visit record, resulting the visit vector $e_*^t \in \mathbb{R}^d$, where $d$ represent the vector dimension and is calculated as follows:

$$e_*^t = W_{*,e} c_*^t \qquad \text{Equation 2}$$

*3.1.2.3. GSVEMed - Graph Visit Representation Learning*

In graph representation learning, the visit vector $e_*^t$ is then passed through a one-layer weighted Graph Convolutional Network (GCN) to become the visit representation $h_*^t$ of each medical code as:

$$h_*^t = \sigma \left( \sum_{j \in \mathcal{N}_t} \alpha_{tj} e_*^t W_* + b \right) \qquad \text{Equation 3}$$

where $\sigma$ is an activation function, $\mathcal{N}_t$ represents the neighbor nodes of node $t$, $W_*$ and $b$ are learnable parameters, and $\alpha$ is attention weight computed using adjacency matrices $A_*$ as:

$$\alpha_{tj} = \frac{exp(A_*[t,j])}{\sum_{k \in \mathcal{N}_t} exp(A_*[t,k])} \qquad \text{Equation 4}$$

$$A_*[t,j] = Jaccard(c_*^i, c_*^j)$$

where $Jaccard(\cdot)$ is Jaccard similarity coefficient.

The nodes in the GCN represent all visit records in the dataset. For each node, only the top $K$ previous visit records with the highest Jaccard similarity scores are connected; here we set *K=10*. This enables visit records with similar medical codes to share similar representation, achieving the objective of "referencing similar past visit records".

### 3.1.2.4. GSVEMed - Aggregated Multi-visits Embedding

Aggregated embedding is the combination of the representations of diagnosis codes, procedure codes, and text representation.

The representation of diagnosis codes $h_d^t$ and procedure codes $h_p^t$, generated from GCN, are initially combined through element-wise addition, then passed through a one-layer Feed Forward Network as Equation 5.

$$v^t = f(h_d^t + h_p^t) \qquad \text{Equation 5}$$

All visit diagnosis codes and procedure codes vectors $V_{d,p}^{(n)} = \left[v_1^{(n)}, v_2^{(n)}, ... v_{T^{(n)}}^{(n)}\right]$ as well as all visit text representations $R^{(n)} = \left[r_1^{(n)}, r_2^{(n)}, ... r_{T^{(n)}}^{(n)}\right]$ are then encoded with their own Transformer Encoder. The output representations of the Transformer Encoder are denoted as $V_{d,p}^{\prime(n)}$ and $R^{\prime(n)}$ respectively. Before the combination, the text representation from Mistral-7B-Instruct-v0.2 has dimensions 4096, whereas the representations of diagnosis and procedure are both in 128 dimensions. Therefore, dimensionality reduction from 4096 to 128 dimensions is necessary for the encoded text representation $R^{\prime(n)}$ through a linear transformation before the addition. After dimensionality reduction, $R^{\prime(n)}$ and the encoded visit code representations $V_{d,p}^{\prime(n)}$ are then aggregated through another element-wise addition as:

$$L^{(n)} = V_{d,p}^{\prime(n)} + R^{\prime(n)} \qquad \text{Equation 6}$$

### 3.1.2.5. GSVEMed - Encoder-Decoder

The aggregated multi-visits representation $V^{(n)}$ is then passed into the Transformer Decoder along with medication vectors $H_m^{(n)} = \left[h_{1,m}^{(n)}, h_{2,m}^{(n)}, ... h_{T^{(n)},m}^{(n)}\right]$, which is shifted-right prior to Equation 7, and output prediction $O^{(n)} = \left[o_1^{(n)}, o_2^{(n)}, ... o_{T^{(n)}}^{(n)}\right]$, where $o_1^{(n)} \in \mathbb{R}^{|\mathcal{C}_m|}$. To ensure that earlier visit records cannot access or reference later ones, both the encoder and decoder use look-ahead masks.

$$O^{(n)} = TransformerDecoder\left(L^{(n)}, H_m^{(n)}\right) \qquad \text{Equation 7}$$

Ultimately, the prediction probabilities of each medication are computed with the sigmoid function as Equation 8.

$$\hat{y}_1, \hat{y}_2, ... \hat{y}_{T^{(n)}} = sigmoid\left(o_1^{(n)}, o_2^{(n)}, ... o_{T^{(n)}}^{(n)}\right) \qquad \text{Equation 8}$$

*3.1.2.6. GSVEMed - Training Objective*

Given that medication recommendation is a multi-label prediction task, we adopt a combination of binary cross-entropy loss $\mathcal{L}_{bce}$ and multi-label margin loss $\mathcal{L}_{multi}$ as our combined loss function. $\mathcal{L}_{multi}$ accounts for the interrelation among multiple labels by quantifying the probability difference between the target class and the not-target class.

$$\mathcal{L}_{bce} = -\sum_t^T \sum_i^{|\mathcal{C}_m|} y_i^t \log \sigma(\hat{y}_i^t) + (1 - y_i^t) \log(1 - \sigma(\hat{y}_i^t)) \qquad \text{Equation 9}$$

$$\mathcal{L}_{multi} = \sum_t^T \sum_i^{|\mathcal{C}_m|} \sum_j^{\hat{Y}^t} \frac{max\left(0, 1 - \left(\hat{y}_t\left[\hat{Y}_j^t\right] - \hat{y}_t[i]\right)\right)}{|\mathcal{C}_m|} \qquad \text{Equation 10}$$

$\mathcal{L}_{bce}$ and $\mathcal{L}_{multi}$ are calculated with Equation 9 and Equation 10 respectively. Here, $\hat{y}_i^t$ and $\hat{y}_t[i]$ represent the predicted probability of element (drug) $i$ in visit $t$, while $\hat{y}_t\left[\hat{Y}_j^t\right]$ is the predicted probability at visit $t$, indexed by the predicted label $j$ in $\hat{Y}_j^t$.

## 3.2. Experimental Setup
## 3.2.1. Dataset Description

The main datasets used in this research are EMR datasets. The sources are MIMIC-III (Johnson et al., 2016) datasets and EMR datasets provided by Ditmanson Medical Foundation Chia-Yi Christian Hospital (CYCH).

### *3.2.1.1. MIMIC-III*

MIMIC-III is a freely accessible dataset released in 2006, comprising data collected from over 40,000 patients who were admitted to critical care units at the Beth Israel Deaconess Medical Center. The data was between 2001 and 2012 and has been de-identified. Including information such as demographics, vital sign measurements made at the bedside, laboratory test results, procedures, medications, caregiver notes, imaging reports, and mortality. The database consists of 26 tables, linked by some identifiers. For example, "SUBJECT_ID" refers to a unique patient, "HADM_ID" is for a unique admission to the hospital, etc.

### *3.2.1.2. CYCH: Ditmanson Medical Foundation Chia-Yi Christian Hospital*

The CYCH dataset cover information from over 20,000 patients hospitalized at CYCH between 2019 and 2022. These datasets contain patients' entire medical history, from admission to discharge, including diagnosis results, medication history, surgery, medical treatments, etc., De-identification has been performed, similar to MIMIC-III. The diagnosis code follows the ICD-10-CM format, the procedure code is formatted with the Medical Service Payment Code of Taiwan's National Health Insurance Administration[1], and the medication codes are formatted according to ATC.

## 3.2.2. Data Preprocessing

Our datasets, MIMIC-III and CYCH, include both structured and unstructured data. The structured data consists of three types of medical codes: diagnosis, procedure, and medication. The unstructured data comprises free-text discharge summary.

---

[1] https://info.nhi.gov.tw/INAE5000/INAE5001S01

*3.2.2.1. Structured Data*

MIMIC-III medical codes are preprocessed following the method of GSVEMed, which adheres the the SafeDrug[3] procedure, but also preserves the patient's records with only one visit. For diagnosis codes, to manage the quantity, only the top 2,000 diagnosis categories are selected, ranked by frequency of occurrence across all diagnosis records. For procedure codes, there are no code removals during the preprocessing. Regarding medication codes, NCD codes undergo conversion to RXCUI format and subsequently to the third-level format. Next, medication categories are filtered based on drug molecule SMILES strings from the DrugBank[4] website.

In CYCH datasets, Medication codes are converted from the ATC format to the third-level format. For the procedure codes, we use the Medical Service Payment Code of Taiwan's National Health Insurance Administration. These codes cover both those used for determining a patient's condition as well as those designated for administrative purposes. Furthermore, some laboratory test item codes from MIMIC-III, including various blood test data, are incorporated. Due to their complexity, physicians at CYCH made a professional judgment to eliminate 223 codes that were not directly pertinent to determining the patient's condition. The handling of diagnosis codes, and the subsequent process is the same as MIMIC-III.

Since GSVEMed learns all visit records in graph visit representation learning and is trained with two or more visit records in the Transformer Encoder-Decoder, the range of covered medical codes may differ between these stages. To maintain consistency in code categories for both MIMIC-III and CYCH, we follow GSVEMed's approach, keeping only the medical codes that are present in both overall patient data and patient records with two or more visits.

*3.2.2.2. Unstructured Data*

In MIMIC-III, most free-text notes are stored in the *NOTEEVENTS* table in English. There are 15 categories of clinical notes in the *NOTEEVENTS* table, we initially filtered only the records that belong to the "Discharge summary" category. Inspired by van Aken et al. (2021)[5], discharge summary are split into admission and discharge summary sections. Notes with the sections of *Chief complaint, Present illness, Medical history, Medications on admission, Allergies, Physical exam, Family history, and Social history* are filtered as admission notes. The discharge summary is the section with the *Procedure, Discharge medication, Discharge diagnosis, Discharge condition, Pertinent results, Hospital course, and Discharge instructions.* The *Discharge medication* section, which contains the medication information and belongs to the predicted objective item, is discarded.

---

[3] https://github.com/ycq091044/SafeDrug
[4] https://go.drugbank.com

As for CYCH, the *discharge summary* and the *admission notes* are two distinct tables for direct access. Most of the notes are in English with some Chinese.

Due to differences in the number of visits between structured and unstructured data, both the MIMIC-III and CYCH discharge summary are filtered to include only the visits present in the structured data. Multiple records of the same visit are concatenated. Extra blanks in the sentence are removed, and the notes are split into chunks to prevent inputting sentences that are too long and trigger out-of-memory errors. We reference the chunking approach outlined by Chang et al. (2023)[6]. This approach is suitable for CYCH with longer sentences. As for MIMIC-III, it is easy to split with the section mentioned in the former. All chunks are guaranteed to be in complete sentences. Visits that only have structured data are all assigned tensor 0 for their text representation. The statistics of the post-processed data are shown in Table 2 (MMIC-III) and Table 3 (CYCH).

---

[5] https://github.com/bvanaken/clinical-outcome-prediction
[6] https://github.com/lilakk/BooookScore

Table 2 Statistics of MMIC-III Dataset

| Item | All Visits | Visits $\geq 2$ |
| --- | --- | --- |
| **Structured Data** | | |
| # of visits / # of patients | 44,107 / 35,425 | 15,032 / 6,350 |
| diag. / prod. / med. space size | 1,958 / 1,430 / 131 | 1,958 / 1,430 / 131 |
| avg. / max # of visits | 1.25 / 29 | 2.37 / 29 |
| avg. / max # of diagnoses per visit | 10.63 / 128 | 10.51 / 128 |
| avg. / max # of procedure per visit | 4.37 / 50 | 3.84 / 50 |
| avg. / max # of medications per visit | 14.91 / 65 | 11.44 / 65 |
| **Unstructured Data** | | |
| # of visits / # of patients | 42,126 / 33,750 | 14,699 / 6,323 |
| avg. length of the discharge summary | 646.81 | 730.15 |
| min / max length of the discharge summary | 15 / 6,209 | 15 / 6,209 |
| avg. # of chunks | 6 | 6 |
| min / max # of chunks | 6 / 6 | 6 / 6 |
| avg. length of each chunk | 107.80 | 121.69 |
| min / max length of chunks | 2 / 5,392 | 2 / 3,758 |

All patients and the edges in the adjacency matrix are split into training, validation, and test sets by the ratio of 4 : 1 : 1. For edges, only edges containing source and target nodes present within the training set are included in the training set. Edges from the training set are also included in both the validation and the test sets. However, edges where both the source and target nodes exist exclusively within the validation set are included solely in the validation set. Likewise, edges where both the source and target nodes exist exclusively within the test set are included solely in the test set.

Table 3 Statistics of CYCH Dataset

| Item | All Visits | Visits $\geq 2$ |
|---|---|---|
| **Structured Data** | | |
| # of visits / # of patients | 52,280 / 22,616 | 38,040 / 8,376 |
| diag. / prod. / med. space size | 1,996 / 1,115 / 122 | 1,996 / 1,115 / 122 |
| avg. / max # of visits | 2.31 / 74 | 4.54 / 74 |
| avg. / max # of diagnoses per visit | 4.03 / 70 | 3.29 / 70 |
| avg. / max # of procedure per visit | 18.11 / 210 | 12.4 / 210 |
| avg. / max # of medications per visit | 5.74 / 61 | 4.45 / 61 |
| **Unstructured Data** | | |
| # of visits / # of patients | 52,229 / 22,601 | 38,004 / 8,376 |
| avg. length of the discharge summary | 1,032.57 | 1,019.97 |
| min / max length of the discharge summary | 62 / 10,299 | 62 / 10,299 |
| avg. # of chunks | 1.3 | 1.27 |
| min / max # of chunks | 1 / 8 | 1 / 8 |
| avg. length of each chunk | 798.44 | 803.13 |
| min / max length of chunks | 1 / 1,977 | 1 / 1,977 |

### 3.2.3. Base Models

We apply our method to the following base models: G-BERT (Shang et al., 2019a), GAMENet (Shang et al., 2019b), SafeDrug (Yang et al., 2021b), COGNet (Wu et al., 2022), GSVEMed (Liao, 2023), SHAPE (Liu et al., 2023) and StratMed (Xiang Li et al., 2024). All of them leverage GNNs in longitudinal data. DAPSNet (Wu et al., 2023a) and MEGACare (Wu et al., 2023b) are not included because the source codes they provided are incomplete, making it difficult to replicate their experiments.

- **G-BERT** was the first to use LM in medication recommendations and incorporated the records of patients with only one visit.

- **GAMENet** incorporated external DDI knowledge and integrated multiple modules.

- **SafeDrug** was the first to incorporate drug molecular structures and enhance the model with the DDI knowledge, differentiating from the approach of GAMENet.

- **COGNet** emphasized the changes in the patient's status, which captured the medication changes among multiple visits.

- **GSVEMed** highlighted referencing similar past visit records representation across patients by using GCN.

- **SHAPE** designed two Encoders: One for visit-level representation and one for patient-level longitudinal representation. They also proposed an adaptive curriculum manager to address the challenge of predicting medication for short visit records that lack historical information.

- **StratMed** emphasized grouping the relationship of the low-frequency data to increase their importance, maintaining performance in both safety and accuracy through dual-property representation constructed by GNN at the visit level.

### 3.2.4. Implementation Setting

The parameters for all base models are set according to the original model's settings in the source code. For GSVEMed, the model learned the graph representation using one GCN layer with a dropout rate of 0.5. Both the Encoder and Decoder consist of one layer, each with the attention head number of 4. The dimension $d$ for medical code embedding is 128, and the loss function weight $\alpha$ is set to 0.95 for MIMIC-III and 0.8 for CYCH. The model is trained for 100 epochs with a batch size of 256, using Adam optimizer and a cosine decay schedule with warmup, and a learning rate of 0.003.

All models use identical training, validation, and test sets during the training stages. Since the original model settings employ different strategies during the inference states, we design them to use the same settings in our experiment. Specifically, G-BERT and GAMENet select the best model from the training checkpoints and perform inference only once. SafeDrug and COGNet select the best model from the training checkpoints but perform inference ten times, each time randomly sampling 80% of the test set. Finally, the results of ten inferences are averaged. Here, we set all the base models to use the average parameters of the top five performance checkpoints, measured using Jaccard Similarity.

For G-BERT, we add additional padding vectors as representations for diagnosis and medication code for predicting the first visit. This is because G-BERT predicts visit $t$ based on the diagnosis and the medication of visit $t-1$, as well as the diagnosis of visit $t$. Nevertheless, the first visit is not included in the computation of the loss function and evaluation metrics. Moreover, to enable training G-BERT on CYCH, we employ the hierarchical division method used by G-BERT for processing ICD-9 ontology to tackle ICD-10-CM[7]. This is due to the diagnosis codes in MIMIC-III are in the ICD-9 format.

### 3.2.5. Evaluation Metrics

#### 3.2.5.1. Jaccard Similarity

The Jaccard Similarity can be used to measure the similarity between two sets of data, revealing shared or distinct items. It is calculated as the intersection of two sets divided by the union of two sets.

$$Jaccard(Y_t, \hat{Y}_t) = \frac{1}{T} \sum_{t=1}^{T} \frac{|Y_t \cap \hat{Y}_t|}{|Y_t \cup \hat{Y}_t|} \quad \text{Equation 11}$$

As shown in Equation 11, $Y_t$ is the ground truth medications and $\hat{Y}_t$ is the medications predicted from the model. The Jaccard Similarity is 0 if the predicted medications do not match any drug with the correct sets and 1 if the two sets are identical.

#### 3.2.5.2. F1-Score

F1-Score combines the precision and recall scores. Precision as Equation 12 measures the proportion of true "positive" medication predictions out of all positive predictions correctly identified by the model, while Recall as Equation 13 represents the proportion of true positive predictions from all actual positive samples in the datasets that were correctly identified by the model. The F1-Score as Equation 14 could simultaneously maximize both precision and recall metrics by using the Harmonic mean. We use macro-averaging to give equal weight to each class.

---
[7] https://www.icd10data.com/ICD10CM/Codes

$$Macro\ Avg.\ Precision = \frac{1}{T}\sum_{t=1}^{T}\frac{|Y_t \cap \hat{Y}_t|}{|\hat{Y}_t|}$$

Equation 12

$$Macro\ Avg.\ Recall = \frac{1}{T}\sum_{t=1}^{T}\frac{|Y_t \cap \hat{Y}_t|}{|Y_t|}$$

Equation 13

$$Macro\ Avg.\ F1 = \frac{1}{T}\sum_{t=1}^{T}\frac{2 \cdot Avg.\ P_t \cdot Avg.\ R_t}{Avg.\ P_t + Avg.\ R_t}$$

Equation 14

### 3.2.5.3. PR-AUC

Precision-Recall Area Under the Curve (PR-AUC) provides a value that summarizes the model's overall performance and is suited for evaluating the precision-recall trade-off. It is useful when comparing the performance of multiple models, allowing us to evaluate our method against the base models using PR-AUC.

$$PR - AUC = \frac{1}{T}\sum_{t=1}^{T}\sum_{k=1}^{|M|} Precision(k)_t \Delta Recall(k)_t$$

Equation 15

$$\Delta Recall(\text{k})_t = Recall(\text{k})_t - Recall(\text{k}-1)_t$$

As shown in Equation 15, $Precision(k)$ and $Recall(k)$ represent the precision and recall at the $k^{th}$ threshold, respectively. $T$ is the total number of visits $t$, $k$ is the rank in the sequence of the retrieved drugs, and $|M|$ is the total number of medications.

### 3.2.5.4. DDI Rate

Drug-drug interactions (DDI) occur when two or more drugs react with each other and could cause the patient to experience some unexpected side effects, while the DDI Rate is used to assess the probability of the interactions. For our experiment, it is defined as the proportion of drug combinations that exhibit interactions out of all possible drug combinations.

$$DDI\ Rate = \frac{\sum_t^T \sum_{i,j} |\{(c_i, c_j) \in \hat{Y}_t | (c_i, c_j) \in A_{DDI}\}|}{\sum_t^T \sum_{i,j} 1} \quad \text{Equation 16}$$

As shown in Equation 16, $(c_i, c_j)$ is the medication pair in predicted medications $\hat{Y}_t$. $A_{DDI}$ is the DDI adjacency matrix, where $A_{DDI}[i, j]$ being 1 if there is an interaction between the $i^{th}$ and the $j^{th}$ medications in the selected items, and 0 if no interactions. $T$ is the total number of visits $t$.

## 3.3. Experimental Design

### 3.3.1. Effectiveness of Medication Recommendation with LLM Text Representation Only

To determine if the extracted knowledge from a general-purpose LLM provides useful information for medication recommendation, we select an LLM that has not been fine-tuned with medical knowledge. The challenge lies in understanding the long and complex free-text context, especially the clinical notes full of medical terminology. Without requiring additional prompts, clinical notes are fed directly into LLM for it to digest in its way. We can then test if the extracted hidden representation captures information relevant for prescribing drugs by evaluating its effectiveness in medication prediction tasks.

### 3.3.2. Combination Representation of Text and Medical Codes

Aside from predicting medication using text representation, we are also interested in exploring whether the medical code embeddings can be appropriately combined with text representations. This involves integrating the structured data with the unstructured data. Similar to the previous sections, we assess the aggregated embeddings for medication prediction.

# 4. Experimental Results

In this chapter, we discuss and analyze the results of the experiments mentioned in Chapter 3.3. Table 4 (MIMIC-III) and Table 5 (CYCH) show overall experiment results. The base model with the notation C represents the original model predicted with medical codes only and the notation T represents the base model predicted with text representation only. However, C+T represents predicting with the combination representation of medical codes and text, where + denotes either addition or concatenation following the original base model design. In CYCH, when the DDI rate control is removed from the loss computation in SafeDrug and StratMed, they are marked as "w/o DDI". "GSVEMed w/ $\alpha = 0.95$" in CYCH is set to compare with GSVEMed in MIMIC-III, where $\alpha$ is set to 0.8. The parameter $\alpha$ controls the weighting of two loss terms, $\mathcal{L}_{bce}$ and $\mathcal{L}_{multi}$.

## 4.1. Performance Comparison

### 4.1.1. Effectiveness of Medication Recommendation with LLM Text Representation Only

We first observe the validity of applying only the text representation to the base models. For MIMIC-III, as shown in Table 4, most of the base models demonstrate a fundamental ability to predict the medication, maintaining a Jaccard Similarity in the forties to fifties. Only COGNet can perform similarly to or even slightly better than the original model using only medical codes.

As for CYCH, as shown in Table 5, G-BERT and GAMENet perform slightly worse than the original model using only medical codes. COGNet and GSVEMed show similar or slightly better performance, while SafeDrug and StratMed struggle to make accurate predictions using only text representation due to the DDI rate control mechanism.

Table 4 Performance Comparison on MIMIC-III. The best results for each base model with different representations are highlighted in bold, and the best overall results among all base models are both bolded and underlined.

| Model | MIMIC-III | | | | | | |
|---|---|---|---|---|---|---|---|
| | Jaccard | F1 | PRAUC | Precision | Recall | DDI | Avg. # of Med. |
| G-BERT$_C$ | 0.5001 | 0.6575 | 0.7675 | 0.5831 | **0.7969** | 0.0845 | 19.08 |
| G-BERT$_T$ | 0.4982 | 0.6562 | 0.7572 | 0.612 | 0.7341 | **0.073** | 13.06 |
| G-BERT$_{C+T}$ | **0.5162** | **0.6721** | **0.776** | **0.6113** | 0.7828 | 0.0869 | 18.15 |
| GAMENet$_C$ | **0.5261** | **0.6807** | 0.7784 | 0.6944 | **0.6949** | 0.086 | 19.48 |
| GAMENet$_T$ | 0.506 | 0.6622 | 0.7645 | 0.7021 | 0.6495 | **0.0793** | 18.01 |
| GAMENet$_{C+T}$ | 0.5252 | 0.6798 | **0.7802** | <u>**0.7149**</u> | 0.6692 | 0.0838 | 18.16 |
| SafeDrug$_C$ | **0.5191** | **0.6752** | **0.772** | **0.6862** | **0.6938** | 0.0647 | 19.49 |
| SafeDrug$_T$ | 0.4412 | 0.6037 | 0.6971 | 0.5824 | 0.6691 | <u>**0.0562**</u> | 21.91 |
| SafeDrug$_{C+T}$ | 0.4426 | 0.6047 | 0.6989 | 0.5769 | 0.678 | 0.0656 | 22.49 |
| COGNet$_C$ | 0.5311 | 0.6845 | 0.7748 | 0.6646 | **0.7392** | 0.0849 | 21.91 |
| COGNet$_T$ | 0.5361 | 0.6893 | 0.7772 | **0.6745** | 0.7324 | 0.0886 | 21.22 |
| COGNet$_{C+T}$ | **0.5369** | **0.6899** | **0.778** | 0.6724 | 0.7366 | 0.0888 | 21.46 |
| GSVEMed$_C$ | 0.5338 | 0.6961 | 0.7724 | **0.6839** | 0.7087 | **0.0832** | 20.4 |
| GSVEMed$_T$ | 0.4581 | 0.6283 | 0.6941 | 0.6665 | 0.5943 | 0.089 | 17.56 |
| GSVEMed$_{C+T}$ | **0.5368** | **0.6986** | **0.7729** | 0.6827 | **0.7153** | 0.0843 | 20.63 |
| SHAPE$_C$ | 0.5456 | 0.6978 | 0.7886 | **0.6901** | 0.7267 | 0.09 | 20.91 |
| SHAPE$_T$ | 0.5192 | 0.6745 | 0.7743 | 0.649 | 0.7293 | 0.0897 | 22.02 |
| SHAPE$_{C+T}$ | <u>**0.5576**</u> | <u>**0.7077**</u> | <u>**0.7996**</u> | 0.6877 | **0.7487** | 0.0855 | 21.57 |
| StratMed$_C$ | 0.5328 | 0.6865 | 0.7766 | **0.6762** | 0.726 | 0.0618 | 20.77 |
| StratMed$_T$ | 0.5014 | 0.658 | 0.7544 | 0.6502 | 0.6968 | **0.0604** | 20.86 |
| StratMed$_{C+T}$ | **0.5329** | **0.6866** | **0.7749** | 0.6747 | **0.7269** | 0.0607 | 20.87 |

C: Medical Codes Only (Original Model); T: Text Representation Only; C+T: Combination of Text and Codes

## 4.1.2. Combination Representation of Text and Medical Codes

After evaluating the effectiveness of the LLM text representation for medication recommendation, we examine the performance of the combined text and medical code representation.

Table 5 Performance Comparison on CYCH. The best results for each base model with different representations are highlighted in bold, and the best overall results among all base models are both bolded and underlined.

| Model | CYCH | | | | | | |
|---|---|---|---|---|---|---|---|
| | Jaccard | F1 | PRAUC | Precision | Recall | DDI | Avg. # of Med. |
| G-BERT$_C$ | 0.4795 | 0.6208 | 0.7124 | 0.5738 | 0.7327 | **0.1282** | 9.51 |
| G-BERT$_T$ | 0.4413 | 0.5897 | 0.6875 | 0.5658 | 0.6596 | 0.1639 | 6.57 |
| G-BERT$_{C+T}$ | **0.4965** | **0.6366** | **0.7311** | **0.5849** | <u>**0.7465**</u> | 0.1332 | 13.06 |
| GAMENet$_C$ | 0.4323 | 0.5739 | 0.6761 | 0.652 | 0.553 | **0.1198** | 9.0903 |
| GAMENet$_T$ | 0.4177 | 0.5587 | 0.6636 | **0.6524** | 0.5263 | 0.1267 | 8.89 |
| GAMENet$_{C+T}$ | **0.4362** | **0.5784** | **0.6791** | 0.6477 | **0.5614** | 0.1202 | 9.33 |
| SafeDrug$_C$ | 0.3216 | 0.471 | 0.5509 | 0.4783 | 0.5079 | **0.0575** | 10.37 |
| SafeDrug$_T$ | 0.2054 | 0.3289 | 0.4105 | 0.4703 | 0.2816 | 0.0667 | 6 |
| SafeDrug$_{C+T}$ | 0.2461 | 0.3826 | 0.4209 | 0.3764 | 0.4463 | 0.0758 | 12 |
| SafeDrug$_C$ w/o DDI | **0.4293** | **0.5792** | **0.6727** | 0.5585 | **0.6451** | 0.1121 | 11.72 |
| SafeDrug$_T$ w/o DDI | 0.2611 | 0.4004 | 0.4398 | 0.4106 | 0.4463 | 0.0909 | 11 |
| SafeDrug$_{C+T}$ w/o DDI | 0.4292 | 0.5783 | 0.6644 | **0.5605** | 0.6363 | 0.1126 | 11.62 |
| COGNet$_C$ | 0.4825 | 0.6189 | 0.6819 | 0.6637 | 0.6224 | 0.1045 | 10.19 |
| COGNet$_T$ | 0.4883 | 0.6241 | 0.6851 | 0.6667 | 0.6272 | 0.1056 | 10.23 |
| COGNet$_{C+T}$ | **0.4913** | **0.6279** | **0.6869** | **0.6686** | **0.6328** | **0.104** | 10.28 |
| GSVEMed$_C$ | 0.5128 | 0.6779 | 0.7446 | 0.6751 | 0.6808 | 0.1067 | 10.75 |
| GSVEMed$_T$ | 0.5166 | 0.6813 | 0.7506 | 0.6695 | **0.6935** | **0.1057** | 11.04 |
| GSVEMed$_{C+T}$ | <u>**0.5292**</u> | <u>**0.6922**</u> | <u>**0.7635**</u> | <u>**0.6946**</u> | 0.6898 | 0.1085 | 10.59 |
| GSVEMed$_C$ w/ $\alpha = 0.95$ | 0.5074 | 0.6732 | 0.7434 | 0.718 | 0.6338 | 0.1154 | 9.41 |
| GSVEMed$_T$ w/ $\alpha = 0.95$ | 0.5115 | 0.6768 | 0.7498 | 0.7187 | 0.6395 | 0.116 | 9.48 |
| GSVEMed$_{C+T}$ w/ $\alpha = 0.95$ | 0.5251 | 0.6886 | 0.7627 | 0.7253 | 0.6554 | 0.1121 | 9.63 |
| SHAPE$_C$ | 0.4886 | 0.6284 | 0.7205 | 0.6559 | **0.6441** | **0.1053** | 10.67 |
| SHAPE$_T$ | 0.4311 | 0.5796 | 0.6769 | 0.5867 | 0.6175 | 0.1106 | 11.22 |
| SHAPE$_{C+T}$ | **0.4901** | **0.6307** | **0.7272** | **0.6707** | 0.6347 | 0.1084 | 10.26 |
| StratMed$_C$ | 0.3372 | 0.4883 | 0.5744 | 0.5182 | 0.5245 | <u>**0.0522**</u> | 10.32 |
| StratMed$_T$ | 0.1626 | 0.2727 | 0.1576 | 0.295 | 0.2919 | 0.1111 | 10 |
| StratMed$_{C+T}$ | 0.1371 | 0.2235 | 0.1621 | 0.3169 | 0.1926 | 0.1333 | 6 |
| StratMed$_C$ w/o DDI | 0.4673 | 0.6118 | 0.693 | **0.5854** | 0.689 | 0.1012 | 12.2 |
| StratMed$_T$ w/o DDI | 0.4 | 0.5489 | 0.6357 | 0.5148 | 0.6395 | 0.1111 | 13.09 |
| StratMed$_{C+T}$ w/o DDI | **0.4673** | **0.612** | **0.6955** | 0.5768 | **0.6987** | 0.0990 | 12.49 |

C: Medical Codes only (Original model); T: Text Representation only; C+T: Combination of Text and Codes

In MIMIC-III, only G-BERT shows an obvious improvement by adding the text representation with the medical code. GAMENet, COGNet, and GSVEMed have a comparable ability to predict with only medical code representation. Only SafeDrug has a limited ability to predict by leveraging the combination representation. In CYCH, G-BERT, COGNet, and GSVEMed obtain visible help by adding additional text representation. GAMENet, SafeDrug, and StratMed still cannot show improvement.

By observing the prediction with combination representation in both MIMIC-III and CYCH, we find that the LLM text representation contains valuable information for the medication recommendation even if not for all base models. The combined representation maintains performance at a certain level and the enhancement of unstructured data utilization implies the presence of valuable information within such data indicating that the LLM text representation does not adversely affect the medical code embeddings or worsen the predictions. These also reveal that the clinical notes in different datasets may contain varying amounts of information.

## 4.2. Results Analysis

The results above illustrate the performance between the two datasets, with only GSVEMed keeping consistent competence.

Notably, DDI rates are generally higher in CYCH compared to MIMIC-III, which can lead to significantly worse outcomes for SafeDrug and StratMed in CYCH. In CYCH, the DDI rates for both SafeDrug and StratMed, whether predicting with text representation alone or with combination representation, exceed the target 0.6 DDI rates in their settings. Early in the training epochs, they had already surpassed 0.6 DDI rates, which caused their predictions to degrade and made it challenging to improve performance. When the DDI rate control is removed in CYCH, they show comparable results with other base models. However, these two models that care about safety and accuracy, SafeDrug and StratMed show varied results in MIMIC-III. StratMed$_T$ and StratMed$_{C+T}$ can maintain its performance while keeping DDI rates low.

Among all the base models, SHAPE is the only one that enhances performance after adding text representation in MIMIC-III but does not show improvement from the text representation in CYCH. Conversely, other models generally perform better after adding text representation in CYCH compared to MIMIC-III. Indicating the varying ability of different models to adapt to different datasets.

# 5. Conclusion

## 5.1. Overall Summary

In this study, we propose a method of enhancing medication recommendation with LLM text representation. We emphasize the effectiveness of LLM text representation in improving medication recommendation and explore its combination with medical codes. By leveraging the language understanding capabilities of off-the-shelf LLM, text representation can be extracted directly from EMR clinical notes without requiring extra preprocessing, pre-training, or fine-tuning. We apply our method to five existing base models and evaluate them on two different datasets MIMIC-III and CYCH. The text representation demonstrates significant progress when combined with the medical codes representation on the G-BERT model in both datasets. Specifically, we observe that with text representation alone, it can achieve comparable ability or potentially superior predictive ability compared to using only medical code representation on COGNet and GSVEMed in the CYCH dataset.

## 5.2. Contributions

To summarize, the main contributions of our work are as follows:

- Leveraging LLM language understanding capabilities to process complex EMR clinical notes without additional preprocessing or training, thereby improving medication recommendation performance.

- Our method is a general method that can be applied to existing base models and shows effectiveness across different datasets.

- By combining text representation with medical codes representation, we enhance the EMR unstructured data utilization and maximize the value of both structured and unstructured data.

## 5.3. Limitations

As stated in section 3.2.2.2, the unstructured clinical notes are too long for LLM to input at once and are split into chunks. This may reduce the completeness of LLM's understanding of the entire single visit record, thereby limiting its effective utilization of contextual understanding capabilities. Inputting the complete clinical notes of a single visit would LLM consider the entire context instead of the separate notes.

When applying text representation to the base models, certain modules or the operations designed for the medical codes representation may not easily adapt to text representation. For instance, the relevance stratification of StratMed maps each medication code with other diagnosis or procedure codes and then updates the embeddings based on edge weight. The edge weights are calculated from the mapping frequency. However, the frequency of the text representation of clinical notes cannot be estimated in the same way as a specific code because each clinical note is mapped with each medication in a visit as a single node. Hence, text representation is directly concatenated with medical codes without additional calculations. This implies that it does not apply the model's designed advantages to the text representation effectively. Therefore, modules for unstructured data still need to be further designed and optimized.

## 5.4. Future Research

We primarily analyze clinical notes such as discharge summary; however, there are still other types of unstructured EMR data, such as admission notes or other semi-structured data. With our method, text representation can further advance progress by incorporating additional representations of these unstructured and semi-structured data.

Moreover, to achieve better integration and understanding of the combination representation of the heterogeneous data, the architectures of the base models may need redesigning, especially in consideration of incorporating text representation from a more complex dimension of LLM.

# Appendix

## A. Combination representation on each base model

### A.1. G-BERT

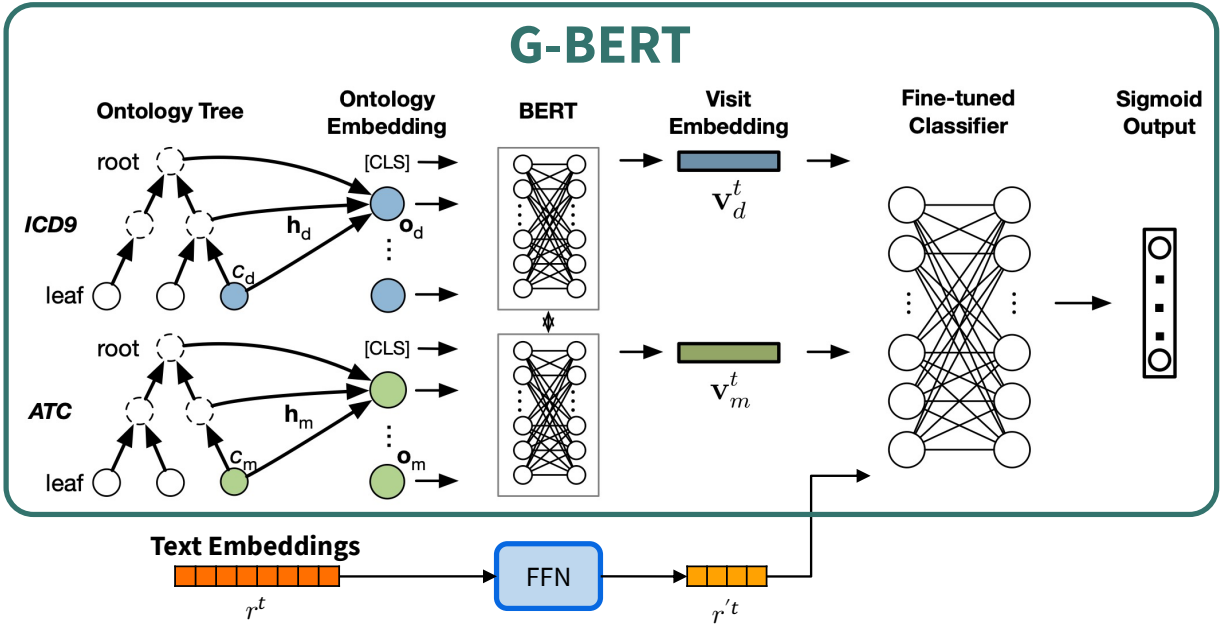

Figure 13 Architecture of G-BERT with LLM Text Representation.

The framework for applying LLM text representation to G-BERT is shown in Figure 13. The text embeddings $r^t$ do not go through pre-training. The prediction layer of the fine-tuning stage is shown in Equation 17.

$$y_t = Sigmoid\left(\boldsymbol{W}_1 \left[\left(\frac{1}{t}\sum_{\mathcal{T}<t} v_d^{\mathcal{T}}\right) \| \left(\frac{1}{t}\sum_{\mathcal{T}<t} v_m^{\mathcal{T}}\right) \| v_d^t \| r'^t\right] + b\right) \qquad \text{Equation 17}$$

where $r'^t$ represents the text embeddings $r^t$ that has been processed through a one-layer Feed Forward Network to reduce their dimension, $\|$ is the concatenate operation.

## A.2. GAMENet

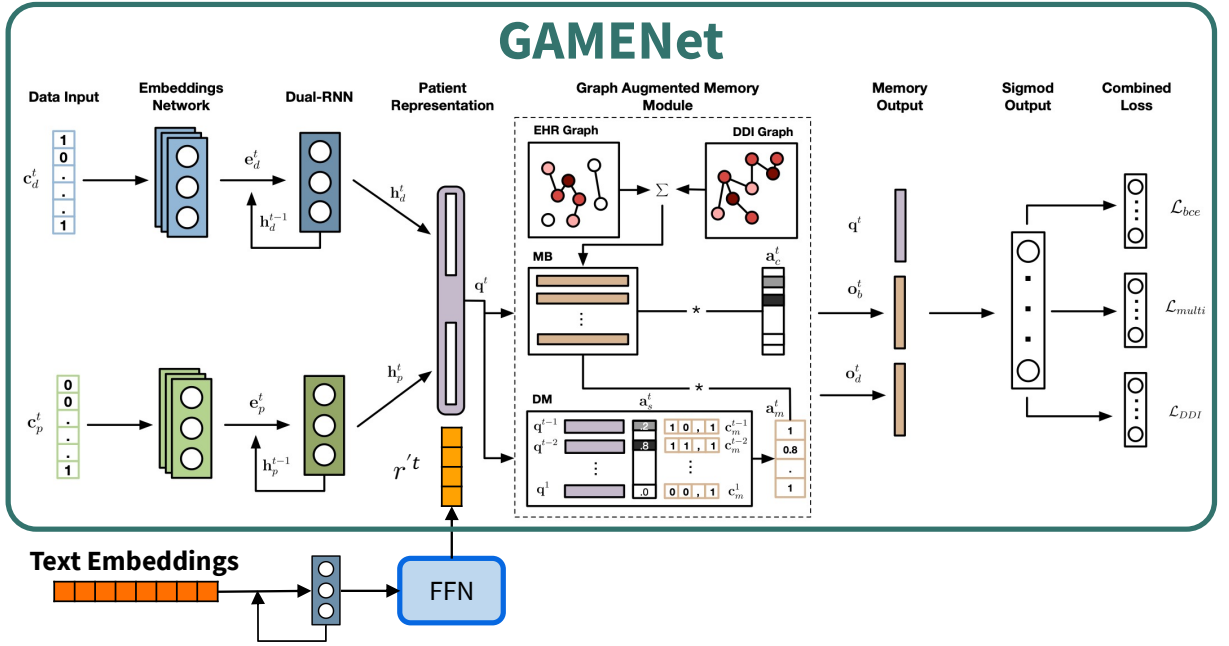

Figure 14 Architecture of GAMENet with LLM Text Representation.

The framework for applying LLM text representation to GAMENet is shown in Figure 14. The input representation to the graph augmented memory module to generate a query is as:

$$q^t = f([h_d^t, h_p^t, r'^t])  \quad \text{Equation 18}$$

where $r'^t$ represents the text embeddings $r^t$ that has been processed through a one-layer Feed Forward Network to reduce their dimension. Then the patient representation $q^t$ and the memory output $o_d^t, o_p^t$ are used to predict the medication as follows:

$$\hat{y}_t = \sigma([q^t, o_d^t, o_p^t])  \quad \text{Equation 19}$$

## A.3. SafeDrug

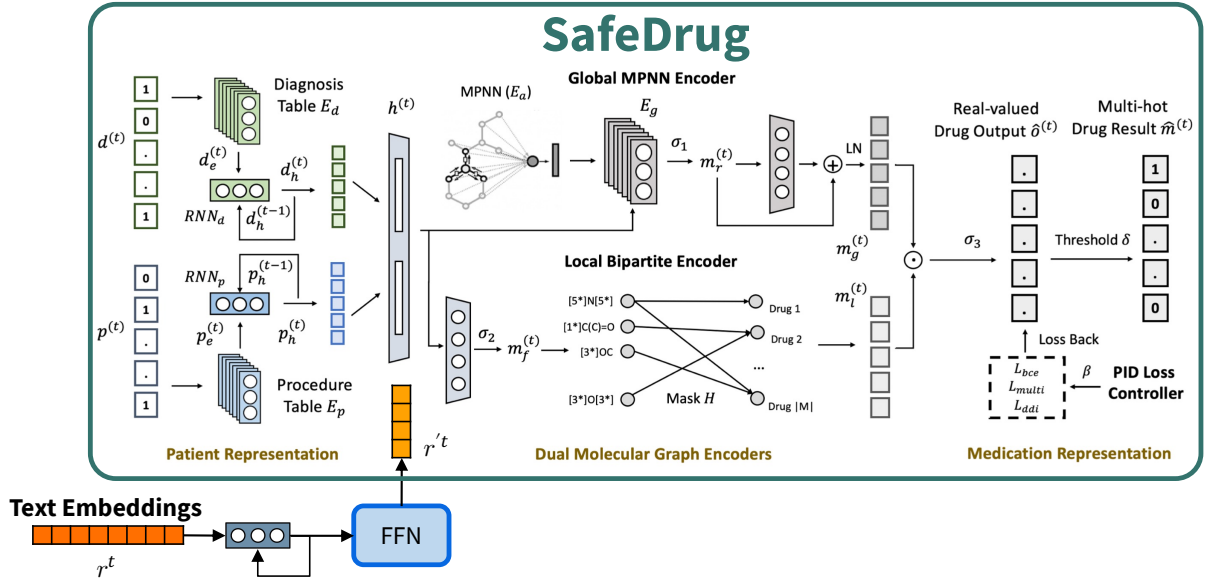

Figure 15 Architecture of SafeDrug with LLM Text Representation.

The framework for applying LLM text representation to SafeDrug is shown in Figure 15. The patient representation $h^{(t)}$ is obtained by concatenating the diagnosis embedding $d_h^{(t)}$, procedure embedding $p_h^{(t)}$, and the text representation $r'^t$ that has undergone dimensional reduction as shown in . After concatenation, the combined representation is passed through a feed-forward network, $\bm{NN}_1(\cdot)$.

$$h^{(t)} = \bm{NN}_1\left(\left[d_h^{(t)}\#p_h^{(t)}\right]; \bm{W}_1\right) \quad \text{Equation 18}$$

After obtained patient representation $h^{(t)}$, the follow-up operations are the same as defined in SafeDrug.

## A.4. COGNet

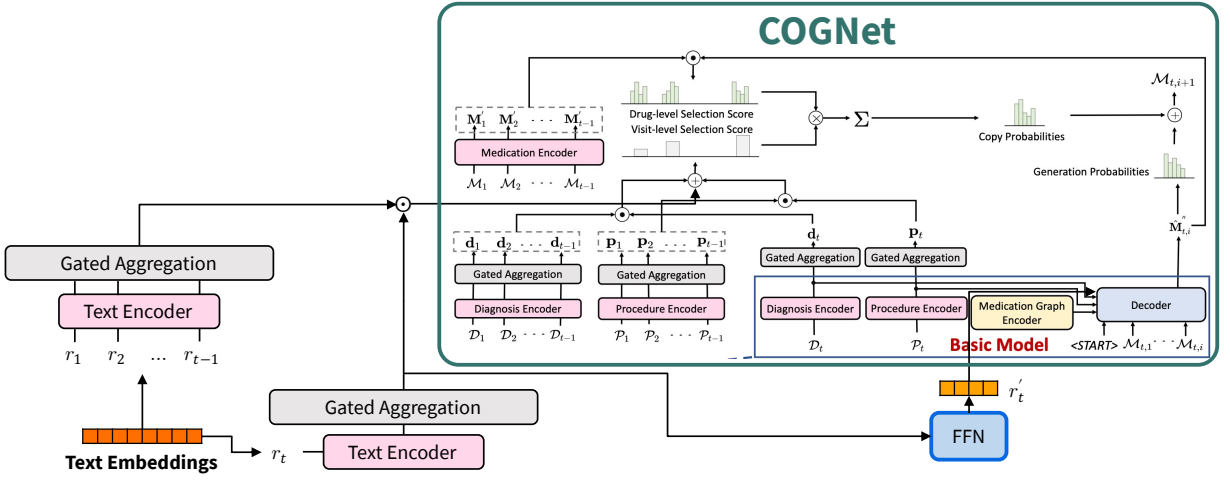

Figure 16 Architecture of COGNet with LLM Text Representation.

The framework for applying LLM text representation to COGNet is shown in Figure 16. In the basic model, the patient's health conditions for Decoder are as follows:

$$\widehat{M}''_t = \text{LayerNorm}\left(\widehat{M}'_t + \text{MH}(\widehat{M}'_t, D'_t, D'_t) + \text{MH}(\widehat{M}'_t, P'_t, P'_t) + MH(\widehat{M}'_t, r'^t, r'^t)\right)$$

Equation 19

then the $i$-th medication is predict via an MLP layer as defined in COGNet:

$$\text{Pr}_g = \text{Softmax}\left(\widehat{M}''_{t,i-1} \mathbf{W}_g + \mathbf{b}_g\right)$$

Equation 20

In the copy module, visit-level health conditions of the text representation are:

$$v_j^r = \text{Softmax}(\tanh(r'^j)r'^j)^\top r'^j$$

Equation 21

then the visit-level selection score of past $j$-th visit are calculate with the $v_j^r$ and the $v_j^d, v_j^p$ defined in COGNet as follows:

$$c_j = \text{Softmax}\left(\frac{v_j^d \cdot v_t^d + v_j^p \cdot v_t^p}{\sqrt{s}}\right) \qquad \text{Equation 22}$$

## A.5. SHAPE

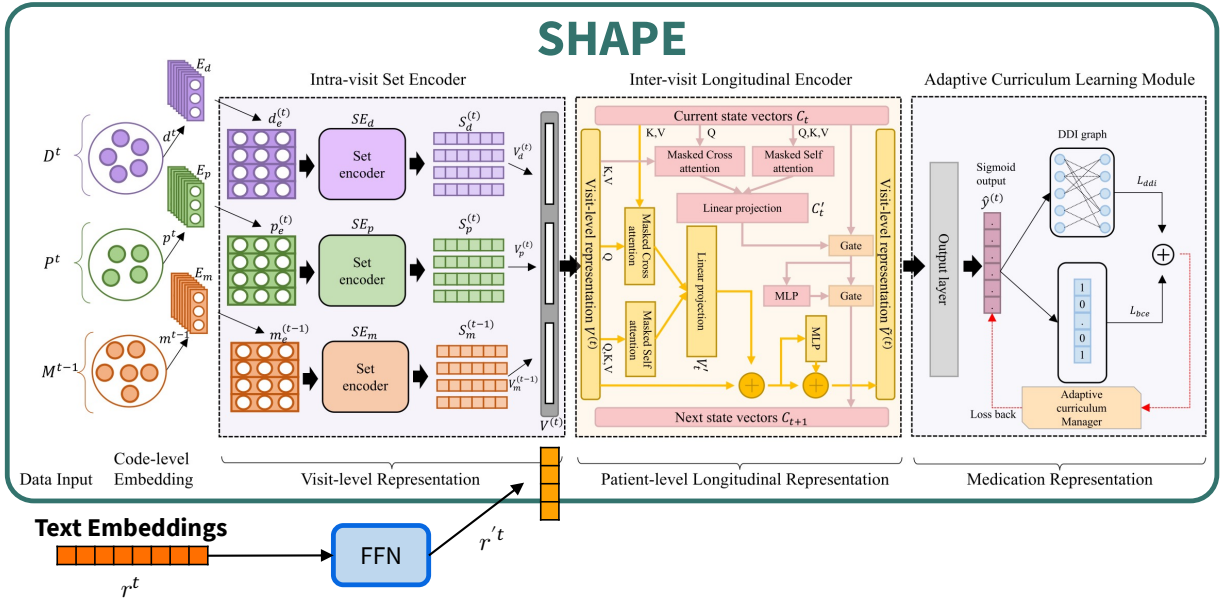

Figure 17 Architecture of SHAPE with LLM Text Representation.

The framework for applying LLM text representation to SHAPE is shown in Figure 17. The visit-level representation $V^{(t)}$ is obtained concatenating the diagnosis code embedding $V_d^t$, procedure code embedding $V_p^t$, medication code embedding $V_m^{(t-1)}$, and the dimensional reduced text representation $r'^t$ as:

$$V^{(t)} = \left[V_d^t, V_p^t, V_m^{(t-1)}\right] \qquad \text{Equation 23}$$

$V^{(t)}$ is then processed through the patient-level longitudinal Encoder and make medication predictions defined in SHAPE.

## A.6. SratMed

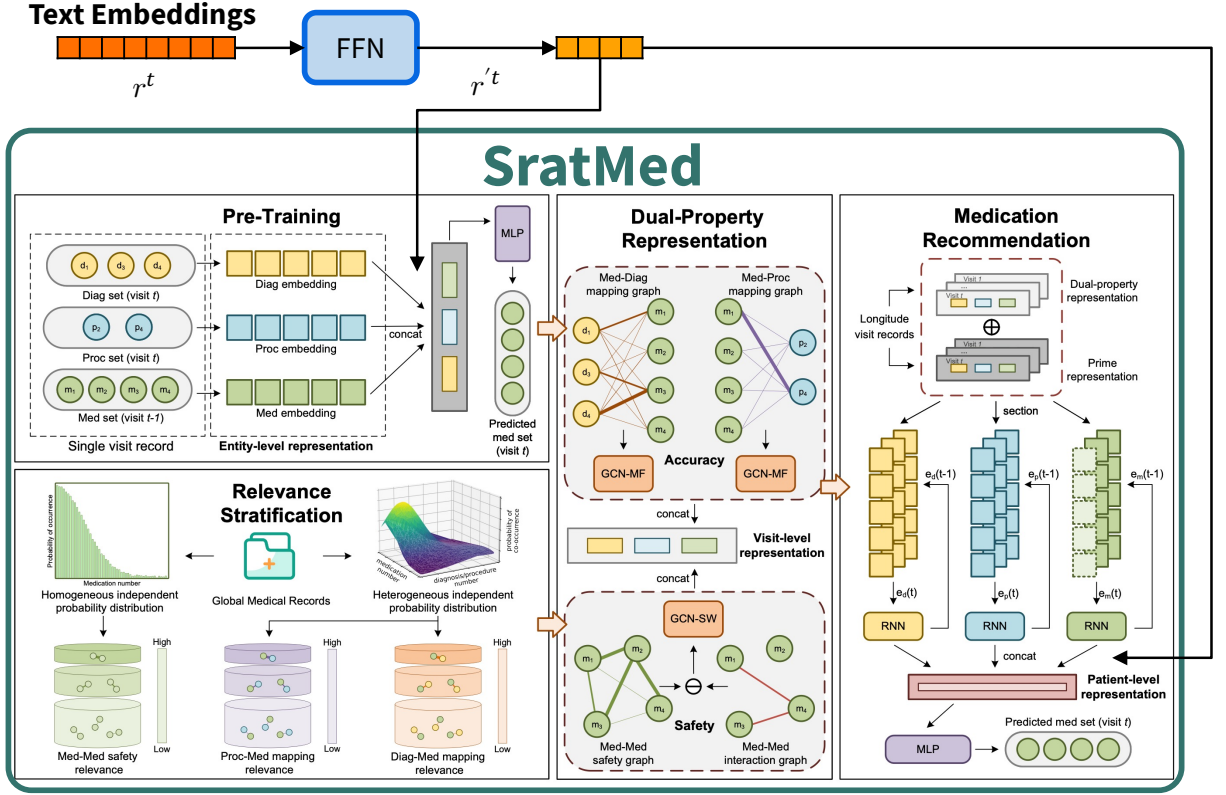

Figure 18 Architecture of StratMed with LLM Text Representation.

The framework for applying LLM text representation to StratMed is shown in Figure 18. In the pre-training stage, the visit-level representation $e'_v(t)$ is concatenated by diagnosis code embedding $e_d(t)$, procedure code embedding $e_p(t)$, medication code embedding $e_m(t-1)$, and the dimensional reduced text representation $r'^t$ as:

$$e'_v(t) = \text{CONCAT}(e_d(t), e_p(t), e_m(t-1), r'^t) \qquad \text{Equation 24}$$

$e'_v(t)$ is then passed through an MLP.

In the training stage, patient-level representation $e^h$ is obtained by concatenating the three RNN-passed medical code representations $[e^h_d, e^h_p, e^h_m]$ as defined by StratMed, along with the text representation $r'^t$, then passed through a MLP as follows:

$$e^h = \text{MLP}\left(\text{CONCAT}(e_d^h, e_p^h, e_m^h, r'^t)\right) \qquad \text{Equation 25}$$

the final medication $\widehat{m}_i$ is also the same as defined in StratMed as:

$$\widehat{m}_i = \begin{cases} 1, & if\ e_i^h \geq \delta \\ 0, & if\ e_i^h < \delta \end{cases} \qquad \text{Equation 26}$$